\documentclass[10pt,twocolumn,letterpaper]{article}

\usepackage{cvpr}              

\usepackage{graphicx}
\usepackage{amsmath}
\usepackage{amssymb}
\usepackage{booktabs}
\usepackage{float} 
\usepackage{algorithm}
\usepackage{algorithmic}
\usepackage{multirow}
\usepackage{tabularx}
\usepackage{wrapfig}
\usepackage[table]{xcolor}
\usepackage{pifont}

\usepackage[pagebackref,breaklinks,colorlinks]{hyperref}
\usepackage[accsupp]{axessibility}  

\usepackage[capitalize]{cleveref}
\crefname{section}{Sec.}{Secs.}
\Crefname{section}{Section}{Sections}
\Crefname{table}{Table}{Tables}
\crefname{table}{Tab.}{Tabs.}

\definecolor{gainsboro}{rgb}{0.66, 0.66, 0.66}

\newcommand{\collapser}[1]{}

\newcommand{\myparagraph}[1]{\vspace{2pt}\noindent{\bf #1}}

\newcommand{\review}[1]{{#1}}

\def\cvprPaperID{2306} 
\def\confName{CVPR}
\def\confYear{2022}

\begin{document}

\title{
CoSSL: Co-Learning of Representation and Classifier for \\Imbalanced Semi-Supervised Learning
}

\author{Yue Fan \qquad Dengxin Dai \qquad Anna Kukleva \qquad Bernt Schiele\\
{\tt\small \{yfan, ddai, akukleva, schiele\}@mpi-inf.mpg.de 
}\\
Max Planck Institute for Informatics, Saarbrücken, Germany
\\ Saarland Informatics Campus
}

\maketitle

\begin{abstract}
Standard semi-supervised learning (SSL) using class-balanced datasets has shown great progress to leverage unlabeled data effectively. However, the more realistic setting of class-imbalanced data -- called imbalanced SSL -- is largely underexplored and standard SSL tends to underperform. 
In this paper, we propose a novel co-learning framework (CoSSL), \review{which decouples representation and classifier learning while coupling them closely.}
To handle the data imbalance, we devise Tail-class Feature Enhancement (TFE) for classifier learning. 
Furthermore, the current evaluation protocol for imbalanced SSL focuses only on balanced test sets, which has limited practicality in real-world scenarios.
Therefore, we further conduct a comprehensive evaluation under various shifted test distributions.
In experiments, we show that our approach outperforms other methods over a large range of shifted distributions, achieving state-of-the-art performance on benchmark datasets ranging from CIFAR-10, CIFAR-100, ImageNet, to Food-101.
Code is available at \url{https://github.com/YUE-FAN/CoSSL}.
\end{abstract}

\section{Introduction} \label{sec:intro}

\begin{figure}[ht]
\centering
\includegraphics[width=0.9\linewidth]{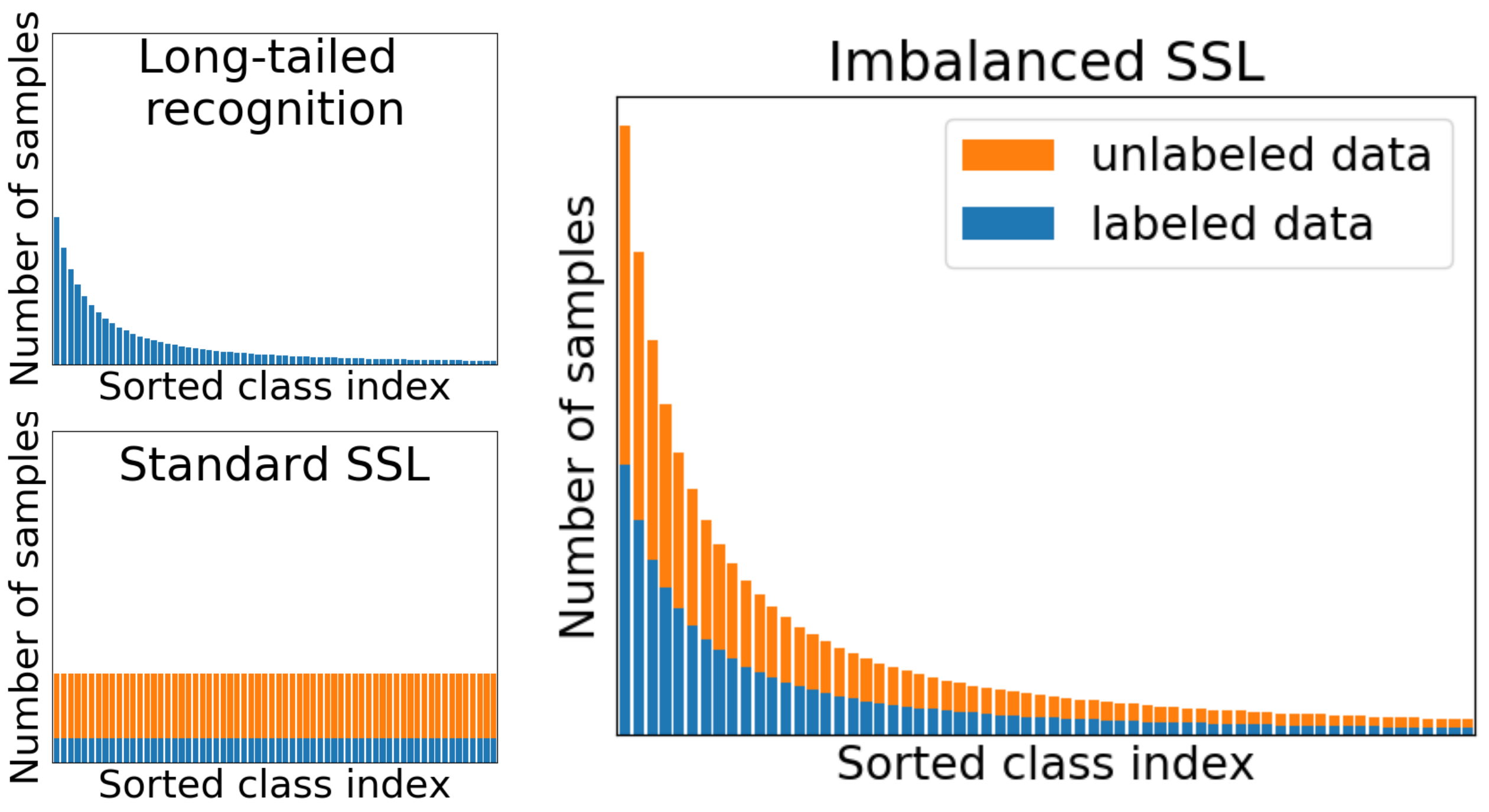}
\caption{
Conventional recognition tasks focus on constrained settings: long-tailed recognition does not involve unlabeled data; semi-supervised learning (SSL) assumes class-balanced distributions for both labeled and unlabeled data.
In this work, we aim at \textit{imbalanced SSL}, where the training data is partially annotated, and both labeled and unlabeled data are not manually balanced.
This setting is more general and poses great challenges to existing algorithms.
A robust learning algorithm should still be able to learn a good classifier under this setting.
}
\label{fig:balanced}
\end{figure}

Imbalanced data distributions are ubiquitous, and pose great challenges for standard deep learning methods.
Many approaches have been proposed 
for long-tailed recognition, where the number of (labeled) examples 
exhibits a long-tailed distribution with heavy class imbalance \cite{liu2019large,guo2016ms,everingham2010pascal,lin2014coco,krishna2017visual,van2017devil}.
While semi-supervised learning (SSL) in the class-balanced setting has shown great promise, in this paper we are interested in the challenging and realistic setting of {\it imbalanced SSL} where both the labeled and the unlabeled data are class-imbalanced, as shown in Fig. \ref{fig:balanced}. 

Despite a few pioneer works \cite{kim2020darp,wei2021crest}, existing solutions from long-tailed recognition and SSL do not generalize well to this setting.
On the one hand, long-tailed recognition \cite{chawla2002smote,he2013imbalanced,he2009learning,huang2016learning,buda2018systematic} is not designed to utilize unlabeled data despite being good at handling data imbalance.
Semi-supervised learning (SSL) \cite{rasmus2015ladder, sajjadi2016firstconsistregular, bachman2014firstconsistregular,scudder1965firstpl,nesterov27firstpl,lee2013pseudo,berthelot2019mixmatch,berthelot2019remixmatch,sohn2020fixmatch}, on the other hand, can effectively leverage unlabeled data but can not address data imbalance.
In certain cases, standard SSL methods trained with imbalanced unlabeled datasets can lead to even worse results than a simple re-balancing method without using any unlabeled data \cite{kim2020darp}, which counters the promise of SSL.

\begin{figure*}[ht]
\centering
\vspace{-7pt}
\includegraphics[width=0.9\textwidth]{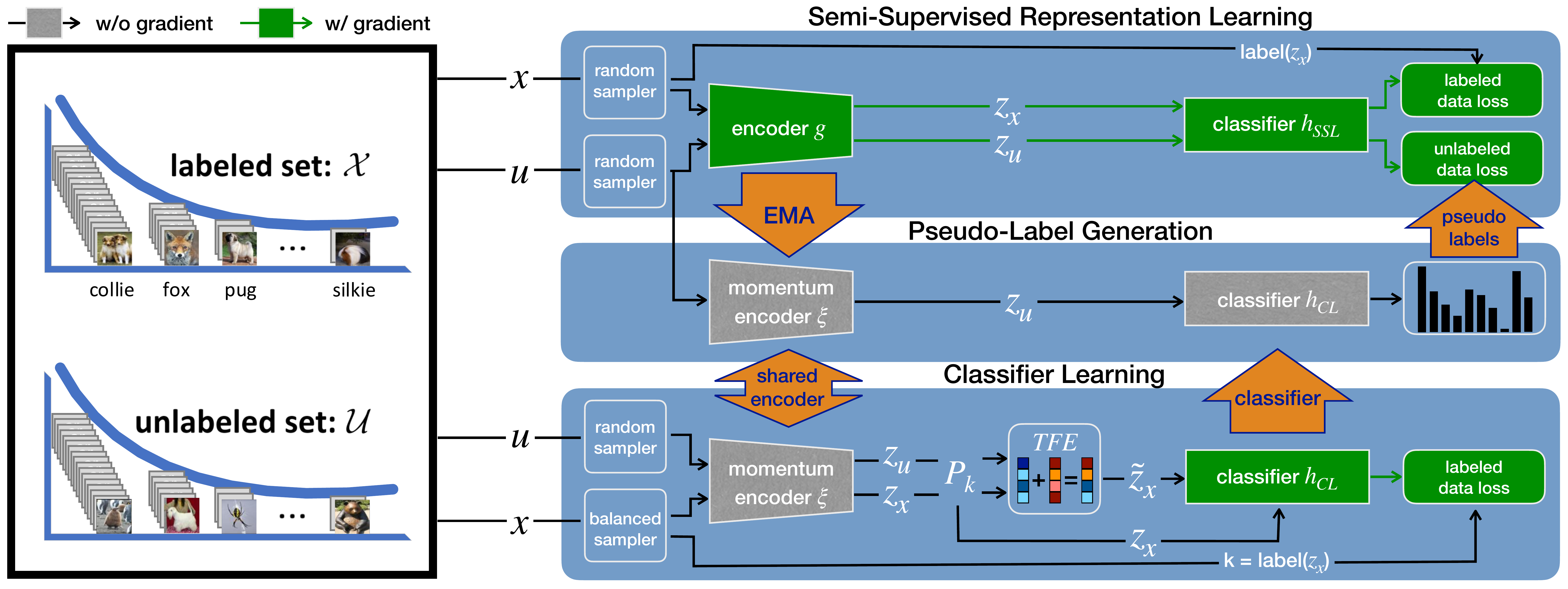}
\caption{
\review{Our co-learning framework CoSSL decouples the training of representation and classifier while coupling them in a non-gradient manner.}
CoSSL consists of three modules: a semi-supervised representation learning module, a balanced classifier learning module, and a carefully designed pseudo-label generation module.
The representation module provides a momentum encoder for feature extraction in the other two modules, and the classifier module produces a balanced classifier using our novel Tail-class Feature Enhancement (TFE).
Then, pseudo-label module generates pseudo-labels for the representation module using the momentum encoder and the balanced classifier.
The interplay between these modules enhances each other, leading to both a more powerful representation and a more balanced classifier.
Additionally, our framework is flexible as it can accommodate any standard SSL methods and classifier learning methods.
}
\label{fig:pipeline}
\vspace{-5pt}
\end{figure*}

\review{
In this paper, we address the imbalanced SSL problem by leveraging strong SSL algorithms \cite{berthelot2019mixmatch,berthelot2019remixmatch,sohn2020fixmatch,xie2019uda} and recent success of decoupling representation and classifier learning from long-tailed recognition \cite{kang2019decouple}.
To this end, we propose CoSSL, a novel co-learning framework for imbalanced SSL, which closely couples representation and classifier while the training of them is decoupled. 
As shown in Fig. \ref{fig:pipeline}, CoSSL consists of three modules: semi-supervised representation learning, classifier learning, and pseudo-label generation.
In our co-learning framework, the representation learning module and the classifier learning module are trained separately without the gradient exchange.
Nonetheless, the two modules in CoSSL are still connected via a shared encoder and pseudo-label generation.
It can then bootstrap itself by exchanging information between the two modules: 1) a shared encoder from the representation learning is passed to classifier training for feature extraction; and 2) the enhanced classifier is used to generate better pseudo-labels for the representation learning.
We show the superiority of our co-learning framework empirically, outperforming previous state-of-the-art methods by a large margin, especially in the case of severe imbalance.
Moreover, we propose Tail-class Feature Enhancement (TFE) for  improved classifier learning for imbalanced SSL, which utilizes unlabeled data as a source of augmentation to enhance the data diversity of tail classes, leading to a more robust classifier.
}

Furthermore, the standard evaluation protocol of long-tailed recognition and SSL normally assumes that the test data are from a uniform class distribution \cite{berthelot2019mixmatch,sohn2020fixmatch,berthelot2019remixmatch,cao2019ldam,kang2019decouple,menon2020long,tang2020long,wang2020devil,li2020overcoming}. 
However, this is insufficient to reflect the diversity of real-world applications, where users may have different needs.
It is strongly desired that the trained model can perform well over a large range of varying distributions, including those that are radically different from the training distribution.
Therefore, in this paper, we adopt the shifted evaluation from \cite{hong2021disentangling}, where the test data are from variously shifted class distributions.
We further distinguish between unknown shifted evaluation and known shifted evaluation, depending on whether test distribution is known a \textit{priori} during training.
This evaluation protocol can be used for long-tailed recognition as well. 

Our contributions are:
(1) We propose a novel co-learning framework CoSSL for imbalanced SSL, which decouples representation and classifier learning while coupling them closely via a shared encoder and pseudo-label generation. 
(2) We devise a novel Tail-class Feature Enhancement (TFE) method to increase the data diversity of tail classes by utilizing unlabeled data, leading to more robust classifiers. 
(3) We propose new evaluation criteria for imbalanced SSL and conduct a comprehensive evaluation.
CoSSL achieves new state-of-the-art results on multiple imbalanced SSL benchmarks across a wide range of evaluation settings.

\section{Related work} \label{sec:related}
\myparagraph{Semi-supervised learning.}
Many efforts have been made in various directions in SSL.
For example, many recent powerful methods \cite{rasmus2015ladder, sajjadi2016firstconsistregular, bachman2014firstconsistregular} are based on consistency regularization, where the idea is that the model should output consistent predictions for perturbed versions of the same input.
Another spectrum of popular approaches is pseudo-labeling~\cite{scudder1965firstpl,nesterov27firstpl,lee2013pseudo} or self-training \cite{rosenberg2005selftrain}, where the model is trained with artificial labels. 
Furthermore, there are many excellent works around generative models \cite{kingma2014deepgan,odena2016gan,denton2016cgan} and graph-based methods \cite{luo2018smoothgraph,liu2019deepgraph,bengio200611graph,joachims2003transductivegraph}. 
A more comprehensive introduction of SSL methods is available in~\cite{chapelle2009semi, zhu05semisurvey, zhu2009semiintro}.
However, none of the aforementioned works have studied SSL in the class-imbalanced setting, in which the standard SSL methods fail to generalize well.

\myparagraph{Long-tailed recognition.}
Research on class-imbalanced supervised learning has
attracted increasing attention.
In particular, many recent efforts have been made to improve the performance under imbalanced data by decoupling the learning of representation and classifier head \cite{kang2019decouple,menon2020long,tang2020long,wang2020devil,li2020overcoming}.
In the two-stage framework from \cite{kang2019decouple}, an instance-balanced sampling scheme was first used for representation learning. 
In the second stage, the classifier head is simply retrained by a class-balanced sampling.
We found that this scheme is also very competitive for imbalanced SSL in our preliminary experiments.
In contrast to this line of works, our co-learning framework focuses on imbalanced SSL and largely simplifies the training pipeline compared to the two-stage framework \cite{kang2019decouple}.
The joint training enables interaction between representation learning and classifier learning, which brings additional benefits to the final performance.
\review{In contrast to BBN \cite{zhou2020bbn}, which has a single loss for two branches and learns the classifier and the representation jointly, CoSSL independently learns classifier and representation with different losses while still connecting them via EMA and pseudo-labeling.}
Evaluation under shifted distributions was also proposed by \cite{hong2021disentangling}, however, we take a step further and consider settings where the test-time distribution is given or not as prior knowledge during the training.

\myparagraph{Imbalanced semi-supervised learning.}
While SSL has been extensively studied, the setting of class-imbalanced semi-supervised is rather under-explored.
Most successful methods from standard SSL do not generalize well to this more realistic scenario without addressing the data imbalance explicitly.
Hyun et al. \cite{hyun2020class} proposed a suppressed consistency loss to suppress the loss on minority classes. 
Kim et al. \cite{kim2020darp} proposed Distribution Aligning Refinery (DARP) to refine raw pseudo-labels via convex optimization.
Wei et al. \cite{wei2021crest} found that the raw SSL methods usually have high recall and low precision for head classes while the reverse is true for the tail classes and further proposed a reverse sampling method for unlabeled data based on that.
\review{
BiS \cite{he2021bis} implements a novel sampler which is helpful for the encoder in the beginning but classifier in the end, however, CoSSL trains the encoder and the classifier independently with different samplers and losses.
In contrast to DASO \cite{oh2021daso}, where pseudo-labels are refined by two complementary classifiers, CoSSL uses a balanced classifier, which is trained by TFE with unlabeled data, to generate pseudo-labels.
Another concurrent work ABC \cite{lee2021abc} introduces an auxiliary classifier which is trained in a balanced way to help the model while sharing the same backbone.
CoSSL differs from ABC \cite{lee2021abc} in: (1) the training of representation and classifier is decoupled; (2) the classifier and the encoder are actively connected to help each other via pseudo-label generation; (3) enhancing tail classes with unlabeled data. 
}

\section{CoSSL: Co-learning for imbalanced SSL} \label{sec:method}

In this section, we first present the problem setup of imbalanced semi-supervised learning (SSL).
Based on this, we introduce CoSSL, a flexible co-learning framework for imbalanced SSL in Section \ref{sec:coteach}.

\myparagraph{Problem setup and notations:}
For a K-class classification problem, there is a labeled set $\mathcal{X}=\{(\textbf{x}_n, y_n): n\in(1,...,N)\}$ and an unlabeled set $\mathcal{U}=\{\textbf{u}_m: m\in(1,...,M)\}$, where $\textbf{x}_n, \textbf{u}_m \in \mathbb{R}^d$ are training examples and $y_n \in \{1,...,K\}$ are class labels for labeled examples.
$N_k$ and $M_k$ denote the numbers of labeled and unlabeled examples in class $k$, respectively, i.e., $\sum_{k=1}^K N_k=N$ and $\sum_{k=1}^K M_k=M$.
Without loss of generality, we assume the classes are sorted by the number of training samples in descending order, i.e., $N_1 \geq N_2 \geq ... \geq N_k$. 
The goal of imbalanced SSL is to train a classifier $f: \mathbb{R}^d \rightarrow \{1,...,K\}$ that generalizes well over a large range of varying test data distributions.

\subsection{Co-learning representation and classifier} \label{sec:coteach}
The two-stage framework \cite{kang2019decouple,menon2020long,tang2020long,wang2020devil,li2020overcoming} from long-tailed recognition is quite successful for supervised learning with imbalanced data.
It decouples representation and classifier by retraining a classifier after the representation learning.
While classifier re-training (cRT) \cite{kang2019decouple} is out-of-the-box a strong baseline, as we will see in the experimental section \ref{sec:main_cifar}, the method has its own limitations when applied to imbalanced SSL: (1) unlabeled data is not utilized during cRT; (2) the two-stage training scheme makes it impossible to refine the pseudo-labels, which in turn limits the quality of feature representation learning.

\review{This motivates us to propose CoSSL, a co-learning framework for imbalanced SSL with a mutual interplay between representations and classifier learning. While decoupling the training of the representations and the classifier, we couple them \textit{without} gradient propagation, so that the final model leverages from the interactions between all the co-modules in our framework. }
As illustrated in Fig. \ref{fig:pipeline}, CoSSL consists of three modules: a semi-supervised representation learning module, a classifier learning module, and a pseudo-label generation module.
The feature encoder from the representation learning module is shared with the classifier module to learn a better classifier, and the improved classifier is used to generate better pseudo-labels for the representation learning module to further improve the feature encoder.
This joint framework largely simplifies the training pipeline compared to the two-stage framework and enables interaction between the representation learning and the classifier learning,
which brings additional benefits to the final performance (see Section \ref{sec:ablation} for ablation).

\myparagraph{Semi-supervised representation learning:}
The goal of the semi-supervised representation learning module is to obtain a strong feature encoder by exploring unlabeled data.
Thanks to the flexibility of our framework, we can use and evaluate a variety of SSL methods \cite{berthelot2019mixmatch,berthelot2019remixmatch,sohn2020fixmatch}.
Given a batch of unlabeled data sampled from the random sampler, we first pass the unlabeled data to the pseudo-label generation module.
Then, the unlabeled data loss is computed using the generated pseudo-labels.
Meanwhile, a batch of labeled data is sampled by the random sampler, and the labeled data loss is computed.
The resulting encoder is accumulated into a momentum encoder and further passed to the classifier module for feature extraction to enhance the classifier training as shown in Fig. \ref{fig:pipeline}.

\myparagraph{Classifier learning with Tail-class Feature Enhancement:}
Inspired by the success of cRT, we train a separate classifier in the classifier learning module and aim to further improve it by using unlabeled data.
To this end, we propose Tail-class Feature Enhancement (TFE) that exploits unlabeled data by blending unlabeled data features with labeled data features while preserving the label of the labeled sample.
Specifically, at each training step, we train the classifier using blended features between labeled and unlabeled data with labels from labeled data.
We deploy a class-balanced sampler and a random sampler to sample a labeled example $(\textbf{x}_i, y_i)$ and an unlabeled example $\textbf{u}_j$.
Then the new fused feature for classifier training is generated by:
\review{
\begin{equation}
  \Tilde{\textbf{z}} = \lambda \xi(\textbf{x}_i) + (1 - \lambda) \xi(\textbf{u}_j) \quad \text{ and } \quad \Tilde{y} = y_i
  \label{eq:feat_mixup_x}
\end{equation}
}
where $\xi$ is the momentum encoder from the representation learning module and the fusion factor $\lambda$ is sampled from a uniform distribution over the interval $[\mu, 1]$. 
\review{We consider samples of $\lambda$ with a value of at least $\mu$ to ensure the validity of the label $y_i$ for the synthesized sample. }

To enhance the data diversity of tail classes, we train the classifier using different portions of fused examples in a stochastic way.
The feature blending is applied with a blend probability that depends on the number of data for each class so that the more labeled data a class has, the less fused data is synthesized for classifier learning.
Formally, given a labeled example from class $k$, we apply feature blending with probability $P_k$ defined as:
\begin{equation}
  P_{k} = \frac{N_1 - N_k}{N_1}
  \label{eq:p_mixup}
\end{equation}
where $N_k$ is the number of examples from the $k$-th class, and $N_1$ is the number of examples of the first class (with the most labeled data). 
Such a class-dependent blend probability encourages more augmented data from feature blending for tail classes, thus, improving the data diversity of tail classes.
For instance, there is no fused data for the first class, which has the most labeled data, since $P_{1} = 0$. 
For a tail class with only 5\% samples of the first class, the blend probability will be as high as 95\%.
Note, that since fused data share the same label with the labeled data, the class distribution is uniform during cRT as the labeled set is sampled using a class-balanced sampler.
Pseudo-code for processing a batch of labeled and unlabeled examples can be found in Alg. \ref{alg}.

\begin{algorithm}[tb]
\footnotesize
\caption{\review{Classifier training with} \textbf{T}ail-class \textbf{F}eature \textbf{E}nhancement}
    \begin{algorithmic}[1]  \label{alg}
    \STATE {\bfseries Input:} Labeled set $\mathcal{X}$, unlabeled set $\mathcal{U}$, feature encoder $\xi$, parameter $\mu$, and batch size $B$ \\
    \FOR{$b = 1$ \TO $B$}
    \STATE \COMMENT{Sample labeled and unlabeled examples} \\
    \STATE $\textbf{x}_i, y_i$ $\sim$ Class-balanced sampler($\mathcal{X}$) \\
    \STATE $\textbf{u}_j$ $\sim$ Random sampler($\mathcal{U}$) \\
    \STATE$P_{y_i} = \frac{N_1 - N_{y_i}}{N_1}$ \COMMENT{Compute the blend probability} \\ 
    \IF{$\text{Uniform}(0, 1) \leq P_{y_i}$ } 
    \STATE \COMMENT{Generate features by feature blending} \\
    \STATE $\lambda \sim \text{Uniform}(\mu, 1)$ \\
    \STATE$\Tilde{\textbf{z}}_b = \lambda \xi(\textbf{x}_i) + (1 - \lambda) \xi(\textbf{u}_j)$ \\
    \STATE$\Tilde{y}_b = y_i$ \\
    \ELSE
    \STATE \COMMENT{Use features of labeled data directly} \\
    \STATE$\Tilde{\textbf{z}}_b = \xi(\textbf{x}_i)$ \\
    \STATE$\Tilde{y}_b = y_i$ \\
    \ENDIF
    \ENDFOR
    \RETURN $\{\Tilde{\textbf{z}}, \Tilde{y}\}$ \COMMENT{Features for classifier training} \\
\end{algorithmic}
\end{algorithm}

\myparagraph{Pseudo-label generation:}
As standard SSL methods suffer from biased pseudo-labels under data imbalance \cite{kim2020darp,wei2021crest}, we devise a pseudo-label generation module to generate high-quality pseudo-labels by combining the strengths of the representation learning module and the classifier learning module.
Given a batch of unlabeled data, it first uses the momentum encoder $\xi$ from the representation learning to extract features since the representations learned from instance-balanced sampling from SSL is the most generalizable \cite{kang2019decouple}.
Then the pseudo-labels are predicted using the classifier trained from TFE leveraging its robustness against data imbalance.
Our pseudo-label generation module replaces the original pseudo-labeling part of the SSL algorithm in the representation learning module and enables the trained classifier to enhance representation learning.
Note, that no gradient updates happen at this step.

\myparagraph{Overall co-learning framework:}
\review{The three aforementioned modules, while being decoupled, are closely coupled with each other in a non-gradient manner.} 
CoSSL can then bootstrap itself by exchanging information between them: the representation learning module provides a momentum encoder for better feature extraction for training classifiers and pseudo-labeling.
And the improved classifier, in turn, generates high-quality pseudo-labels to further enhance representation learning.
Specifically, denote the overall network by $f$, which consists of a feature extractor network $g(\cdot)$ and a classifier head $h(\cdot)$. 
At training iteration $t$, the three modules operate successively as shown in Fig. \ref{fig:pipeline}.
(1) For the classifier module, a batch of labeled data and unlabeled data from $\mathcal{X}$ and $\mathcal{U}$ are sampled using a class-balanced sampler and a random sampler,
respectively.
Then, the features are extracted by a momentum encoder $\xi(\cdot)$ of $g(\cdot)$, which is provided by the representation learning module.
We update $\xi$ by $\xi_{t} = m \xi_{t-1} + (1 - m) g_t$ where $\xi_0=g_0$ and $m \in [0, 1)$ is a momentum coefficient.
Then, the classifier $h$ is trained using our TFE with standard cross-entropy loss.
(2) For the pseudo-label generation module, it encodes a new batch of unlabeled data with the same momentum encoder $\xi_{t}$ and predicts the pseudo-labels using the classifier $h$ from the classifier module.
(3) The generated pseudo-labels are then fed into the representation module to compute the unlabeled data loss.
Meanwhile, a new batch of labeled data is used in the representation module.

CoSSL fits particularly well for imbalanced SSL as the representation module and the classifier module, despite being decoupled, can enhance each other via pseudo-labeling and the momentum encoder, leading to both a more powerful representation and a more balanced classifier.
\review{We find empirically that coupling representation and classifier without explicit gradient propagation leads to a better performance than variants with it. (see Section \ref{sec:ablation}).}
Moreover, our co-learning framework is very flexible as it can accommodate any standard SSL algorithm and classifier learning method, which makes it possible to benefit from the most advanced approaches.
We present the complete algorithm of our co-learning framework in the Appendix.

\section{Experimental evaluation}
In this section, we conduct extensive experiments to evaluate the efficacy of our framework.
In Section \ref{sec:main_cifar}, \ref{sec:main_small127}, and \ref{sec:main_food}, we compare our method with existing works and show that we achieve state-of-the-art performance for the commonly used uniform test evaluation.
Section \ref{sec:skewed} evaluates different methods over a large range of imbalance settings, and we distinguish between two cases: the distributions are unknown or known a priori during training.
A detailed analysis of our framework can be found in Section \ref{sec:ablation}.

\subsection{Main results on CIFAR-10 and CIFAR-100} \label{sec:main_cifar}

\myparagraph{Datasets.}
Following common practice \cite{cui2019class,cao2019ldam}, we employ CIFAR10-LT and CIFAR100-LT for imbalanced SSL by randomly selecting some training images for each class determined by a pre-defined imbalance ratio $\gamma$ as the labeled and the unlabeled set.
Specifically, we set $N_k = N_1 \cdot \gamma^{-\frac{k-1}{K-1}}$ for labeled data and $M_k = M_1 \cdot \gamma^{-\frac{k-1}{K-1}}$ for unlabeled data.
\review{For results in the main paper,} we use $N_1 = 1500$; $M_1 = 3000$ for CIFAR-10 and $N_1 = 150$; $M_1 = 300$ for CIFAR-100, respectively. 
Following \cite{kim2020darp,wei2021crest}, we report results with imbalance ratio $\gamma=50$, 100 and 150 for CIFAR10-LT and $\gamma=20$, 50 and 100 for CIFAR100-LT.
Therefore, the number of labeled samples for the least class is 10 and 1 for CIFAR-10 with $\gamma=150$ and CIFAR-100 with $\gamma=100$, respectively.
\review{Results on more settings can be found in Appendix.}

\myparagraph{Setup.}
Following \cite{kim2020darp,cirecsan2010deep}, we evaluate our method with MixMatch~\cite{berthelot2019mixmatch}, ReMixMatch~\cite{berthelot2019remixmatch}, and FixMatch~\cite{sohn2020fixmatch} under the same implementation (as recommended by \cite{oliver2018realistic}) using Wide ResNet-28-2 \cite{zagoruyko2016wrn} as the backbone.
The hyper-parameter $\mu$ in Alg. \ref{alg} is set to $0.6$ based on the ablation study in Section \ref{sec:ablation}.
We apply TFE module in the last 20\% of iterations for faster training and better accuracy (see Appendix for more details).
As our implementation is based on the public codebase from \cite{kim2020darp}, we use the same hyper-parameters as theirs.
For example, all experiments are trained with batch size 64 using Adam optimizer \cite{kingma2014adam} with a constant learning rate of 0.002 without any decay.
We train all models for 500 epochs, each of which has 500 steps, resulting in a total number of $2.5 \times 10^5$ training iterations.
For all experiments, we report the average test accuracy of the last 20 epochs following \cite{oliver2018realistic}.
For CReST+, we use the official TensorFlow implementation.
As for data augmentation for TFE, we use the strong augmentation from \cite{sohn2020fixmatch}, which consists of RandAugment \cite{cubuk2020randaugment} and CutOut \cite{devries2017cutout}.

\begin{table}[]
\centering
\resizebox{\linewidth}{!}{%
\begin{tabular}{llll}
    \toprule
     & \multicolumn{3}{c}{CIFAR-10-LT} \\ \cmidrule(l{3pt}r{3pt}){2-4}
     & \multicolumn{1}{c}{$\gamma$=50} & \multicolumn{1}{c}{$\gamma$=100} & \multicolumn{1}{c}{$\gamma$=150} \\ \cmidrule(l{3pt}r{3pt}){1-1} \cmidrule(l{3pt}r{3pt}){2-4}
    vanilla & $65.2_{\pm0.05}^*$ & $58.8_{\pm0.13}^*$ & $55.6_{\pm0.43}^*$ \\ \cmidrule(l{3pt}r{3pt}){1-1} \cmidrule(l{3pt}r{3pt}){2-4}
    \multicolumn{4}{c}{Long-tailed recognition methods} \\ \cmidrule(l{3pt}r{3pt}){1-1} \cmidrule(l{3pt}r{3pt}){2-4}
    w/ Re-sampling \cite{japkowicz2000resample} & $64.3_{\pm0.48}^*$ & $55.8_{\pm0.47}^*$ & $52.2_{\pm0.05}^*$ \\
    w/ LDAM-DRW \cite{cao2019ldam} & $68.9_{\pm0.07}^*$ & $62.8_{\pm0.17}^*$ & $57.9_{\pm0.20}^*$ \\
    w/ cRT \cite{kang2019decouple} & $67.8_{\pm0.13}^*$ & $63.2_{\pm0.45}^*$ & $59.3_{\pm0.10}^*$ \\ \cmidrule(l{3pt}r{3pt}){1-1} \cmidrule(l{3pt}r{3pt}){2-4}
    \multicolumn{4}{c}{SSL methods} \\ \cmidrule(l{3pt}r{3pt}){1-1} \cmidrule(l{3pt}r{3pt}){2-4}
    MixMatch \cite{berthelot2019mixmatch} & $73.2_{\pm0.56}^*$ & $64.8_{\pm0.28}^*$ & $62.5_{\pm0.31}^*$ \\
    w/ DARP \cite{kim2020darp} & $75.2_{\pm0.47}^*$ & $67.9_{\pm0.14}^*$ & $65.8_{\pm0.52}^*$ \\
    w/ CReST+ \cite{wei2021crest} & $79.0_{\pm0.26}^*$ & $71.9_{\pm0.33}^*$ & $68.3_{\pm0.57}^*$ \\
    w/ CoSSL & $\textbf{80.31}_{\pm0.31}$ & $\textbf{76.4}_{\pm1.14}$ & $\textbf{73.5}_{\pm1.25}$ \\ \cmidrule(l{3pt}r{3pt}){1-1} \cmidrule(l{3pt}r{3pt}){2-4}
    ReMixMatch \cite{berthelot2019remixmatch} & $81.5_{\pm0.26}^*$ & $73.8_{\pm0.38}^*$ & $69.9_{\pm0.47}^*$ \\
    \review{w/ Re-sampling} \cite{japkowicz2000resample} & $83.6_{\pm0.54}$ & $76.7_{\pm0.24}$ & $71.5_{\pm0.64}$ \\
    \review{w/ LDAM-DRW} \cite{cao2019ldam} & $85.9_{\pm0.23}$ & $80.5_{\pm0.71}$ & $76.1_{\pm0.53}$ \\
    w/ DARP \cite{kim2020darp} & $82.1_{\pm0.14}^*$ & $75.8_{\pm0.09}^*$ & $71.0_{\pm0.27}^*$ \\
    \review{w/ DARP + cRT} \cite{kim2020darp} & $87.3_{\pm0.16}^*$ & $83.5_{\pm0.07}^*$ & $79.7_{\pm0.54}^*$ \\
    w/ CReST+ \cite{wei2021crest} & $83.7_{\pm0.15}$ & $78.8_{\pm0.54}$ & $75.2_{\pm0.30}$ \\
    \review{w/ CReST+ + LA} \cite{wei2021crest} & $84.2_{\pm0.11}$ & $81.3_{\pm0.34}$ & $79.2_{\pm0.31}$ \\
    w/ CoSSL & $\textbf{87.7}_{\pm0.21}$ & $\textbf{84.1}_{\pm0.56}$ & $\textbf{81.3}_{\pm0.83}$ \\ \cmidrule(l{3pt}r{3pt}){1-1} \cmidrule(l{3pt}r{3pt}){2-4}
    FixMatch \cite{sohn2020fixmatch} & $79.2_{\pm0.33}^*$ & $71.5_{\pm0.72}^*$ & $68.4_{\pm0.15}^*$ \\
    \review{w/ Re-sampling} \cite{japkowicz2000resample} & $84.8_{\pm0.21}$ & $78.9_{\pm0.63}$ & $75.2_{\pm0.45}$ \\
    \review{w/ LDAM-DRW} \cite{cao2019ldam} & $80.0_{\pm0.60}$ & $73.1_{\pm0.81}$ & $69.1_{\pm0.51}$ \\
    w/ DARP \cite{kim2020darp} & $81.8_{\pm0.24}^*$ & $75.5_{\pm0.04}^*$ & $70.4_{\pm0.25}^*$ \\
    \review{w/ DARP + cRT}\cite{kim2020darp} & $85.8_{\pm0.43}$ & $82.4_{\pm0.26}$ & $79.6_{\pm0.42}$ \\
    w/ CReST+ \cite{wei2021crest} & $83.9_{\pm0.14}^*$ & $77.4_{\pm0.36}^*$ & $72.8_{\pm0.58}^*$ \\
    \review{w/ CReST+ + LA} \cite{wei2021crest} & $84.9_{\pm0.02}$ & $80.8_{\pm0.20}$ & $77.5_{\pm0.74}$ \\
    w/ CoSSL & $\textbf{86.8}_{\pm0.30}$ & $\textbf{83.2}_{\pm0.49}$ & $\textbf{80.3}_{\pm0.55}$ \\ \bottomrule
\end{tabular}
}%
\caption{
Classification accuracy (\%) on CIFAR-10-LT using a Wide ResNet-28-2 under the uniform test distribution of three different class-imbalance ratios $\gamma$.
The numbers are averaged over 5 different folds.
We use the same code base as \cite{kim2020darp} for fair comparison following \cite{oliver2018realistic}.
Numbers with $^*$ are taken from the original papers.
The best number is in bold.
}
\label{tab:cifar10}
\vspace{-15pt}
\end{table}

\begin{table}[]
\centering
\resizebox{\linewidth}{!}{%
\begin{tabular}{llll}
    \toprule
     & \multicolumn{3}{c}{CIFAR-100-LT} \\ \cmidrule(l{3pt}r{3pt}){2-4}
     & \multicolumn{1}{c}{$\gamma$=20} & \multicolumn{1}{c}{$\gamma$=50} & \multicolumn{1}{c}{$\gamma$=100} \\ \cmidrule(l{3pt}r{3pt}){1-1} \cmidrule(l{3pt}r{3pt}){2-4}
    ReMixMatch \cite{berthelot2019remixmatch} & $51.6_{\pm0.43}$ & $44.2_{\pm0.59}$ & $39.3_{\pm0.43}$ \\
    \review{w/ Re-sampling} \cite{japkowicz2000resample} & $50.0_{\pm0.56}$ & $42.9_{\pm0.95}$ & $37.8_{\pm0.46}$ \\
    \review{w/ LDAM-DRW} \cite{cao2019ldam} & $54.5_{\pm0.95}$ & $47.5_{\pm0.79}$ & $42.3_{\pm0.35}$ \\
    w/ DARP \cite{kim2020darp} & $51.9_{\pm0.35}$ & $44.7_{\pm0.66}$ & $39.8_{\pm0.53}$ \\
    \review{w/ DARP + cRT} \cite{kim2020darp} & $54.5_{\pm0.42}$ & $48.5_{\pm0.91}$ & $43.7_{\pm0.81}$ \\
    w/ CReST+ \cite{wei2021crest} & $51.3_{\pm0.34}$ & $45.5_{\pm0.76}$ & $41.0_{\pm0.78}$ \\
    \review{w/ CReST+ + LA} \cite{wei2021crest} & $51.9_{\pm0.60}$ & $46.6_{\pm1.14}$ & $41.7_{\pm0.69}$ \\
    w/ CoSSL & $\textbf{55.8}_{\pm0.62}$ & $\textbf{48.9}_{\pm0.61}$ & $\textbf{44.1}_{\pm0.59}$ \\ \cmidrule(l{3pt}r{3pt}){1-1} \cmidrule(l{3pt}r{3pt}){2-4}
    FixMatch \cite{sohn2020fixmatch} & $49.6_{\pm0.78}$ & $42.1_{\pm0.33}$ & $37.6_{\pm0.48}$ \\
    \review{w/ Re-sampling} \cite{japkowicz2000resample} & $49.9_{\pm0.76}$ & $43.2_{\pm0.54}$ & $38.2_{\pm0.60}$ \\
    \review{w/ LDAM-DRW} \cite{cao2019ldam} & $51.1_{\pm0.45}$ & $40.4_{\pm0.46}$ & $34.7_{\pm0.22}$ \\
    w/ DARP \cite{kim2020darp} & $50.8_{\pm0.77}$ & $43.1_{\pm0.54}$ & $38.3_{\pm0.47}$ \\
    \review{w/ DARP + cRT} \cite{kim2020darp} & $51.4_{\pm0.68}$ & $44.9_{\pm0.54}$ & $40.4_{\pm0.78}$ \\
    w/ CReST+ \cite{wei2021crest} & $51.8_{\pm0.12}$ & $44.9_{\pm0.50}$ & $40.1_{\pm0.65}$ \\
    \review{w/ CReST+ + LA} \cite{wei2021crest} & $52.9_{\pm0.07}$ & $47.3_{\pm0.17}$ & $42.7_{\pm0.70}$ \\
    w/ CoSSL & $\textbf{53.9}_{\pm0.78}$ & $\textbf{47.6}_{\pm0.57}$ & $\textbf{43.0}_{\pm0.61}$ \\ \bottomrule
\end{tabular}
}%
\caption{
Classification accuracy (\%) on CIFAR-100-LT under the uniform test distribution of three different class-imbalance ratios $\gamma$.
The numbers are averaged over 5 different folds.
We reproduce all numbers using the same codebase from \cite{kim2020darp} for a fair comparison\protect\footnotemark.
The best number is in bold.
}
\vspace{-10pt}
\label{tab:cifar100}
\end{table}

\myparagraph{Results.}
Table \ref{tab:cifar10} and Table \ref{tab:cifar100} compare our method with various SSL algorithms and long-tailed recognition algorithms on CIFAR-10-LT and CIFAR-100-LT with various imbalance ratios $\gamma$.
Our method achieves the best performance across all settings with significant margins over the previous state-of-the-art.
Noticeably, our method is particularly good at larger imbalance ratios. For example, we outperform the second-best by an absolute accuracy of $7.5\%$ on CIFAR-10-LT at imbalance ratio $\gamma = 150$ with FixMatch, which underlines the superiority of our method.
Replacing MixMatch with ReMixMatch or FixMatch as the representation learning module can increase test accuracy on CIFAR-10-LT at imbalance ratio $\gamma=150$ by $7.8\%$ and $6.8\%$, respectively.
On CIFAR-100-LT, we evaluate our method on top of ReMixMatch and FixMatch as they give the best performance on CIFAR-10-LT.
Besides the best performance across settings, our method also improves performance for small imbalance ratios as well ($4.5\%$ higher than the second-best at imbalance ratio $\gamma=20$ with ReMixMatch).

\footnotetext{Note that the results from \cite{kim2020darp} with $\gamma=20$ are not used here because they were produced by $N_1=300, M_1=150$: \url{https://github.com/bbuing9/DARP/blob/master/run.sh}}

\subsection{Main results on Small-ImageNet-127} \label{sec:main_small127}

\myparagraph{Dataset.}
ImageNet127 is originally introduced in \cite{huh2016makes} and used by \cite{wei2021crest} for imbalanced SSL.
It is a naturally imbalanced dataset with imbalance ratio $\gamma \approx 286$ by grouping the 1000 classes of ImageNet \cite{deng2009imagenet} 
into 127 classes based on the WordNet hierarchy. 
Due to limited resources, we are not able to conduct experiments on ImageNet127 with the full resolution\protect\footnotemark.
Instead, we propose a down-sampled version of ImageNet127 to test the effectiveness of our method on a large-scale dataset.
Inspired by \cite{chrabaszcz2017small}, we down-sample the original images from ImageNet127 to smaller images of $32\times32$ or $64\times64$ pixels using the box method from Pillow library (different down-sampling techniques yield very similar performance as pointed out by \cite{chrabaszcz2017small}).
Following \cite{wei2021crest}, we randomly select 10\% training samples as the labeled set.
The test set is unchanged, and averaged class recall is used to achieve a balanced metric.

\footnotetext{One run of vanilla FixMatch on ImageNet127 on a single NVIDIA Tesla V100 takes 10676.5 hours which is about 444 days.}

\myparagraph{Setup and results.}
We evaluate our method using FixMatch \cite{sohn2020fixmatch} with ResNet-50 \cite{he2016resnet} due to its good performance on CIFAR.
For all experiments, we train for a total number of 500 epochs.
For CReST+, we train for 5 generations with 100 epoch per generation.
The rest of hyper-parameters are the same as used in CIFAR-LT.
\review{As for data augmentation of TFE, we use random crop and horizontal flipping.
Table \ref{tab:small127food101} summarizes the results on Small-ImageNet-127.
CoSSL achieves the best and the second-best performance for image sizes 32 and 64, respectively.}

\begin{table}[ht]
\resizebox{\linewidth}{!}{%
\begin{tabular}{lcccc}
\toprule
\multirow{2}{*}{} & \multicolumn{2}{c}{Small-ImageNet-127} & \multicolumn{2}{c}{Food-101-LT} \\ \cmidrule(l{3pt}r{3pt}){2-3} \cmidrule(l{3pt}r{3pt}){4-5}
 & $32 \times 32$ & $64 \times 64$ & $\gamma = 50$ & $\gamma = 100$ \\ \cmidrule(l{3pt}r{3pt}){1-1} \cmidrule(l{3pt}r{3pt}){2-5}
FixMatch &  29.7 & 42.3  & 42.6 & 35.3 \\
w/ DARP \cite{kim2020darp} &  30.5 & 42.5 & 42.0 & 34.2 \\
\review{w/ DARP + cRT} \cite{kim2020darp} & 39.7 & 51.0 & 41.5 & 34.4 \\
w/ CReST+ \cite{wei2021crest} & 32.5 & 44.7 & 43.8 & 31.2 \\
\review{w/ CReST+ + LA} \cite{wei2021crest} & 40.9 & \textbf{55.9} & 47.7 & 36.1 \\
w/ CoSSL & \review{\textbf{43.7}} & \review{53.8} & \review{\textbf{49.0}} & \review{\textbf{40.4}} \\ \bottomrule
\end{tabular}
}%
\caption{
Averaged class recall (\%) on Small-ImageNet-127 and Food-101.
We test image size $32\times32$ and $64\times64$ for Small-ImageNet-127 and $\gamma = 50$ and $\gamma = 100$ for Food-101. 
}
\label{tab:small127food101}
\vspace{-5pt}
\end{table}

\begin{table*}[ht]
\resizebox{\textwidth}{!}{%
\centering
\begin{tabular}{lccccccccccccccccccccc}
\toprule
\multirow{2}{*}{Test imbalance ratio} & 512 & 256 & \textbf{150} & 128 & 64 & 32 & 16 & 8 & 4 & \multicolumn{1}{c|}{2} & \multicolumn{1}{c|}{1} & -2 & -4 & -8 & -16 & -32 & -64 & -128 & -256 & -512 & Mean \\  \cmidrule(l{3pt}r{3pt}){2-21} \cmidrule(l{3pt}r{3pt}){22-22}
 & \multicolumn{21}{c}{\cellcolor{blue!25}\textbf{Unknown test-time imbalance ratio}} \\ \cmidrule(l{3pt}r{3pt}){1-1} \cmidrule(l{3pt}r{3pt}){2-21} \cmidrule(l{3pt}r{3pt}){22-22}
Fix & 94.83 & 93.95 & 93.13 & 92.87 & 91.24 & 89.11 & 86.62 & 82.90 & 78.92 & \multicolumn{1}{c|}{73.58} & \multicolumn{1}{c|}{67.83} & 61.83 & 55.41 & 49.50 & 44.46 & 40.37 & 36.88 & 33.89 & 30.95 & 29.04 & 66.36 \\ \cmidrule(l{3pt}r{3pt}){1-1} \cmidrule(l{3pt}r{3pt}){2-21} \cmidrule(l{3pt}r{3pt}){22-22}
Fix + PC & 94.63 & 93.95 & 93.30 & 92.95 & 91.54 & 89.89 & 87.87 & 84.89 & 82.05 & \multicolumn{1}{c|}{77.97} & \multicolumn{1}{c|}{73.49} & 68.86 & 63.88 & 59.45 & 55.70 & 52.76 & 50.24 & 47.90 & 45.77 & 44.23 & 72.57 \\
Fix + vanilla cRT & 94.78 & 93.90 & 93.17 & 92.83 & 91.24 & 89.24 & 86.87 & 83.75 & 80.29 & \multicolumn{1}{c|}{75.54} & \multicolumn{1}{c|}{70.40} & 65.10 & 59.47 & 54.36 & 49.86 & 46.35 & 43.39 & 40.81 & 38.34 & 36.61 & 69.31 \\ \cmidrule(l{3pt}r{3pt}){1-1} \cmidrule(l{3pt}r{3pt}){2-21} \cmidrule(l{3pt}r{3pt}){22-22}
Fix + DARP & \textbf{95.14} & \textbf{94.46} & \textbf{93.73} & \textbf{93.50} & \textbf{92.18} & \textbf{90.12} & 87.70 & 84.39 & 81.03 & \multicolumn{1}{c|}{76.26} & \multicolumn{1}{c|}{71.15} & 66.12 & 60.99 & 56.10 & 52.28 & 48.84 & 45.75 & 43.25 & 40.79 & 39.17 & 70.65 \\
Fix + CReST+ & 94.18 & 93.39 & 92.74 & 92.45 & 91.05 & 89.04 & 86.70 & 83.52 & 80.20 & \multicolumn{1}{c|}{76.05} & \multicolumn{1}{c|}{71.75} & 67.28 & 62.76 & 58.73 & 55.68 & 52.89 & 50.47 & 48.49 & 46.61 & 45.54 & 71.98 \\
Fix + CoSSL & 91.73 & 91.13 & 90.90 & 90.60 & 89.85 & 89.07 & \textbf{87.95} & \textbf{86.24} & \textbf{84.60} & \multicolumn{1}{c|}{\textbf{82.61}} & \multicolumn{1}{c|}{\textbf{80.40}} & \textbf{78.39} & \textbf{76.03} & \textbf{74.19} & \textbf{73.21} & \textbf{72.49} & \textbf{71.43} & \textbf{70.64} & \textbf{70.02} & \textbf{69.71} & \textbf{81.06} \\ \cmidrule(l{3pt}r{3pt}){1-1} \cmidrule(l{3pt}r{3pt}){2-21} \cmidrule(l{3pt}r{3pt}){22-22}
 & \multicolumn{21}{c}{\cellcolor{blue!25}\textbf{Known test-time imbalance ratio}} \\ \cmidrule(l{3pt}r{3pt}){1-1} \cmidrule(l{3pt}r{3pt}){2-21} \cmidrule(l{3pt}r{3pt}){22-22}
Fix + PC & 94.98 & 94.00 & 93.13 & 92.83 & 91.16 & 89.24 & 87.03 & 84.00 & 81.03 & \multicolumn{1}{c|}{77.31} & \multicolumn{1}{c|}{73.49} & 70.10 & 66.79 & 64.21 & 62.69 & 61.89 & 62.41 & 63.26 & 64.80 & 66.50 & 77.04 \\
Fix + vanilla cRT & 95.14 & 94.32 & 93.39 & 93.25 & 91.35 & 89.24 & 86.73 & 83.45 & 79.85 & \multicolumn{1}{c|}{75.04} & \multicolumn{1}{c|}{70.40} & 65.76 & 60.65 & 56.67 & 53.81 & 52.04 & 51.07 & 51.09 & 49.98 & 51.60 & 72.24 \\
Fix + DARP + PC & \textbf{95.19} & \textbf{94.46} & \textbf{93.73} &\textbf{ 93.54} & \textbf{92.32} & \textbf{90.32} & \textbf{88.17} & \textbf{85.53} & 83.00 & \multicolumn{1}{c|}{79.96} & \multicolumn{1}{c|}{76.82} & 74.33 & 72.05 & 70.88 & 70.37 & 70.53 & 70.98 & 71.39 & 72.19 & 73.07 & 80.94 \\
Fix + CReST+ + PC & 94.48 & 93.44 & 92.74 & 92.49 & 91.09 & 89.17 & 87.20 & 84.75 & 82.60 & \multicolumn{1}{c|}{79.86} & \multicolumn{1}{c|}{77.74} & 76.09 & 74.41 & 74.03 & 74.40 & 75.40 & 76.38 & 77.22 & 78.66 & 80.29 & 82.62 \\
Fix + CoSSL + PC & 92.83 & 91.59 & 90.90 & 90.31 & 89.22 & 87.93 & 86.42 & 85.01 & \textbf{84.00} & \multicolumn{1}{c|}{\textbf{82.57}} & \multicolumn{1}{c|}{\textbf{82.00}} &\textbf{ 81.70} & \textbf{81.72} & \textbf{81.66} & \textbf{82.94} & \textbf{84.66} & \textbf{85.77} & \textbf{86.83} & \textbf{87.58} & \textbf{88.31} & \textbf{86.20} \\
\bottomrule
\end{tabular}
}%
\caption{
Classification accuracy (\%) on CIFAR-10-LT with imbalance ratio $\gamma=150$.
We test different methods on top of FixMatch \cite{sohn2020fixmatch} for known and unknown shifted distributions.
Post-compensation (PC) \cite{hong2021disentangling} is deployed to utilize the information of the known test distribution. 
}
\vspace{-5pt}
\label{tab:skewed}
\end{table*}

\subsection{Main results on Food-101} \label{sec:main_food}

\myparagraph{Dataset.}
To evaluate the effectiveness of our method on high-resolution images, we use the fine-grained image classification dataset Food-101 \cite{bossard2014food101}.
The original dataset consists of 101 food categories, with 101,000 images. For each class, 250 manually reviewed test images are provided as well as 750 training images.
All images were rescaled to have a maximum side length of 512 pixels.
We construct Food-101-LT for imbalanced SSL using the same way as CIFAR-10-LT with imbalance ratio $\gamma=50$ and $100$.

\myparagraph{Setup.}
We consider FixMatch \cite{sohn2020fixmatch} as the SSL algorithm due to its good performance.
We train a ResNet-50 \cite{he2016resnet} for 1,000 epochs of unlabeled dataset using a SGD optimizer with momentum 0.9.
The learning rate is set to 0.04 without decay, with a linear warm-up for the first 5 epochs.
We set the labeled batch size as 256 and the unlabeled batch size as 512.
The EMA decay rate is 0.999.
\review{We use random crop and horizontal flipping for TFE.}

\myparagraph{Results.}
Table \ref{tab:small127food101} shows the results on Food-101-LT. 
Compared to other methods, which give marginal improvements or, in some cases, even worse performance over the baseline, our method consistently improves the accuracy.
\review{We outperform the second-best by $1.3\%$ and $4.3\%$ at imbalance ratio $\gamma=50$ and 100, respectively.}

\subsection{Evaluation at unknown and known shifted test distributions} \label{sec:skewed}
As mentioned above, the standard evaluation under uniform test distribution is often limited in reflecting real-world scenarios.
To this end, we conduct a more realistic evaluation by assessing different methods at shifted test distributions.
Moreover, we argue that the test distribution can be given as prior knowledge in real-world applications in some cases.
Thus, we distinguish two types of shifted evaluation: known test distributions in which the test distribution is given during training, and unknown test distributions in which this information is unknown. 
When the test distribution is known, the imbalanced SSL method should be able to accommodate the information for further improvement.

Inspired by \cite{hong2021disentangling}, we construct shifted test sets with a wide range of imbalance ratios.
When $\gamma > 0$, the number of test examples of class $k$ is defined as $N_k = N_1 \cdot \gamma^{-\frac{k-1}{K-1}}$, where class 1 has the most test data.
Similarly, $N_k = N_1 \cdot |\gamma|^{\frac{k-1}{K-1}}$ when $\gamma < 0$, where class 1 has the least test data, and, thus, test set is weighted in favor of tail classes.
For unknown distributions, we train different methods and evaluate them directly over a family of shifted distributions.
The mean accuracy is also reported. 
When the distribution is known during training, we deploy post-compensation \cite{hong2021disentangling} as a post-processing method to utilize this information for all methods.
For all experiments, we use FixMatch and train on CIFAR-10-LT with imbalance ratio $\gamma=150$.
Then, we evaluate different methods at unknown and known shifted test distributions varying from imbalance ratio $\gamma=512$ to $-512$. 
All experiments are run with the same data split and the training protocol from Section \ref{sec:main_cifar}.
Results of other training settings can be found in the Appendix.

Table \ref{tab:skewed} summarizes the results.
Compared to other methods, our approach has higher mean accuracy for both known and unknown distributions, which is mainly due to the good performance at the negative test imbalance ratios.
For example, while being lower at positive ratios, our method is $24.17\%$ and $8.02\%$ better than the second-best at imbalance ratio $\gamma=-512$ in known and unknown cases, respectively.
Our method also shows good robustness against the change of test imbalance ratios.
For known test distribution, as the information of test distributions is utilized during the training in our method, we achieve a more balanced performance under various imbalance ratios. 
For example, the performance gap between $\gamma=512$ and $\gamma=-512$ is $4.52\%$ for our method compared to $14.19\%$ for CReST+ and $22.12\%$ for DARP.
Despite the improved performance from our method, the relatively lower results at the negative ratios also indicate that none of the existing methods, including ours, can achieve a real balanced performance.
Note that our protocol can be applied for imbalanced supervised learning as well.

\begin{table*}[!ht]
    \footnotesize{
    \begin{subtable}[!h]{0.2\textwidth}
        \centering
        \begin{tabular}{cc}
        \toprule
        allow grad & 76.46 \\
        CoSSL & \textbf{80.24} \\ \bottomrule
        \end{tabular}
        \caption{Benefits of stop-gradient}
        \begin{tabular}{cc}
        \toprule
        $h_{SSL}$ & 78.23 \\
        $h_{CL}$ & \textbf{80.24} \\ \bottomrule
        \end{tabular}
        \caption{Pseudo-label generation}
    \end{subtable}
    \begin{subtable}[!h]{0.2\textwidth}
        \centering
        \begin{tabular}{cc}
        \toprule
        - & 69.16 \\
        two-stage & 73.52 \\
        CoSSL & \textbf{80.24} \\ \bottomrule
        \end{tabular}
        \caption{Benefits of decoupling}
    \end{subtable}
    \begin{subtable}[!h]{0.3\textwidth}
        \centering
        \begin{tabular}{cc}
        \toprule
         - & 77.22 \\
         mixUp\cite{zhang2017mixup} & 77.36 \\
         MFW\cite{ye2021procrustean} & 77.91 \\
         TFE & \textbf{80.24} \\ 
         \bottomrule
        \end{tabular}
        \caption{Different classifier learning methods in CoSSL}
    \end{subtable}
    \begin{subtable}[!h]{0.3\textwidth}
        \centering
        \begin{tabular}{lc}
        \toprule
        blend labels with pseudo-labels & 73.24 \\
        image-level enhancement & 78.92 \\
        remove blend probability & 77.90 \\ \midrule
        TFE & \textbf{80.24} \\ \bottomrule
        \end{tabular}
        \caption{Design choices in TFE}
    \end{subtable}
    }
    \caption{
    \textbf{(a)} Performance degrades if representation is updated with gradients from the classifier module.
    \textbf{(b)} Benefits of using classifier learning module to generate pseudo-labels.
    $h_{SSL}$ and $h_{CL}$ are classifiers from the representation learning and the classifier learning module, respectively.
    \textbf{(c)} Both decoupled approaches (two-stage, CoSSL) show better results over the joint training. Particularly, our co-learning achieves the best performance across settings.
    \textbf{(d)} Test accuracy of different classifier learning methods in CoSSL.
    \textbf{(e)} Design choices in TFE.
    }
    \vspace{-5pt}
    \label{tab:cossl_ablation}
\end{table*}

\subsection{Ablation study} \label{sec:ablation}
\review{
In this section, we first analyze different design choices for CoSSL to provide additional insights into how it helps generalization.
Then, we provide detailed ablation studies on TFE.
We use CIFAR-10-LT with $\gamma=150$ as our main ablation settings.
We focus on a single split and report results for a Wide ResNet-28-2 \cite{zagoruyko2016wrn} with FixMatch \cite{sohn2020fixmatch} backbone.
For fair comparison, the same data split is used for all experiments in this section.
Ablation on other settings can be found in Appendix.
}

\myparagraph{Benefits of the co-learning framework.} \label{sec:cossl_ablation}
\review{
We attribute the success of CoSSL to four aspects.
(1) Decoupling representation and classifier is crucial for imbalanced SSL, and our co-learning framework which further couples them closely is superior to the standard two-stage approach.
As is shown in Table \ref{tab:cossl_ablation} (c), both decoupled training schemes (co-learning and two-stage) show significant performance improvement over the joint training method.
In particular, our co-learning approach CoSSL shows preferred test accuracy to the two-stage approach across settings, which suggests the importance of coupling representation and classifier while being decoupled.
(2) It is more beneficial to not update the representation directly with the gradient from the classifier learning module.
In Table \ref{tab:cossl_ablation} (a), test accuracy shows 3.78\% drop when representation is updated with gradients from the classifier learning module.
(3) Instead, it is advantageous to use the balanced classifier $h_{CL}$ for pseudo-label generation due to its robustness against data imbalance, as is shown in Table \ref{tab:cossl_ablation} (b). 
(4) Last but not least, it is important to utilize unlabeled data for classifier learning, and modifications we proposed in TFE are important for the final performance.
Table \ref{tab:cossl_ablation} (d) compares performance of different classifier learning strategies for CoSSL.
Methods that leverage unlabeled data (MFW\cite{ye2021procrustean} and TFE) outperform the ones that do not (cRT and cRT+) in most cases.
In particular, TFE achieves the best accuracy across different settings, which justifies its importance to CoSSL.
}

\myparagraph{Design choices in TFE.} \label{sec:tfe_ablation}
\review{
Dedicated to imbalanced SSL, TFE differs from existing feature mixing approaches in three important aspects.
First, we utilize class-dependent blend probability $P_k$ to encourage more augmentation for the tail classes, thus, improving the final performance as is shown in Table \ref{tab:cossl_ablation} (e).
Removing the mechanism of $P_k$ decreases the performance by 2.34\%.
Second, the fusion factor $\lambda$ is sampled from a uniform distribution between $\mu$ and 1.
}
\review{
This strategy shows better empirical results than the commonly used beta distribution and other variants of uniform distribution (see Appendix).
Thirdly, TFE does not apply label blending. 
Table \ref{tab:cossl_ablation} (e) shows a performance drop of 7.00\% when labels are mixed with pseudo-labels from unlabeled data.
TFE does not only show the best performance in our joint framework but also shows the best performance in the two-stage framework (see Appendix D).
}

\section{Conclusion and limitations} \label{sec:conclusion}
In this work, we study imbalanced SSL, which is a more general setting as both labeled and unlabeled data from imbalanced distributions.
We propose CoSSL, a flexible co-learning framework for imbalanced SSL, which decouples the representation learning and classifier learning while connecting them by sharing learned features and generated pseudo-labels.
We also design Tail-class Feature Enhancement for learning the classifier with unlabeled data and enhancing the performance at tail classes.
Integrating TFE and strong SSL methods into our CoSSL framework, we achieve new state-of-the-art results across a variety of imbalanced SSL benchmarks, especially when the imbalance ratio is large.
At the evaluation, we address the limitation of the conventional uniform protocol by evaluating methods at shifted distributions and considering known and unknown test distribution during training. 
Such a comprehensive evaluation provides more insights into the existing methods and uncovers limitations.

This work, however, is also subject to several limitations.
First, this paper focuses on the object recognition problem under class-imbalanced distribution.
Therefore, caution must be taken when generalizing to other vision tasks.
Second, our method only considers in-class unlabeled data whose potential class labels are covered by the labeled set.
However, there are often a large number of out-of-class unlabeled data available in real-world applications. 
And they are often mixed with in-class unlabeled data, which can be detrimental if not properly handled.
Our method, at the current stage, is not able to handle such a case and effectively leverage out-of-class unlabeled data, which we leave for future work.
Thirdly, as we have seen from Section \ref{sec:skewed}, all of the existing methods, including ours, can not achieve a real balanced performance across test distributions.
The performance at distributions that are radically different from the training distribution is relatively lower.

{\small
\bibliographystyle{ieee_fullname}
\bibliography{egbib}
}

\newpage\hbox{}\thispagestyle{empty}\newpage
\appendix

In the supplementary material, we first present results for two additional experimental setups in Section \ref{sec:rlru} and \ref{sec:lessdata}.
Then, we provide pseudo-code of our co-learning framework CoSSL in Section \ref{sec:pseudocode}.
More ablation studies of our co-learning framework can be found in Section \ref{sec:ablation}.
Finally, we conduct more evaluation at unknown and known shifted test distributions in Section \ref{sec:skewed}.

\section{Evaluation under $\gamma_l \neq \gamma_u$} \label{sec:rlru}
The imbalance ratio of labeled data is not always the same as that of unlabeled data in practice.
In this section, we compare different methods under $\gamma_l \neq \gamma_u$.
Table \ref{tab:rlru} shows the results on CIFAR-100 with $\gamma_l=50$ and $\gamma_u=100$ with two different SSL backbone.
In both cases, CoSSL gives superior performance to other methods.

\begin{table}[!ht]
\centering
\begin{tabular}{lc}
\toprule
CIFAR-100 & $\gamma_l=50$ $\gamma_l=100$ \\ \midrule
ReMixMatch & 42.07 \\
w/ DARP & 43.19 \\
w/ DARP + cRT & 46.59 \\
w/ CReST+ & 42.31 \\
w/ CReST+ + LA & 41.42 \\
w/ CoSSL & \textbf{47.33} \\ \midrule
FixMatch & 40.47 \\
w/ DARP & 41.20 \\
w/ DARP + cRT & 43.01 \\
w/ CReST+ & 41.20 \\
w/ CReST+ + LA & 44.14 \\
w/ CoSSL & \textbf{45.92} \\ \bottomrule
\end{tabular}
\caption{
Comparison on CIFAR-100 with $\gamma_l=50$, $\gamma_u=100$.
}
\label{tab:rlru}
\end{table}

\section{Evaluation with less number of labeled data} \label{sec:lessdata}
In this section, we provide more evaluation of our method with less number of labeled data than that of the main paper.
We compare different methods on CIFAR-100 with $\gamma=100$.
We set the number of labeled data as 50 for the first class.
As is shown in Table \ref{tab:less}, CoSSL outperforms other methods and achieves the best performance.

\begin{table}[!ht]
\centering
\resizebox{0.48\linewidth}{!}{%
\begin{tabular}{lcc}
\hline
CIFAR-100 & $N_1$=50 \\ \hline
ReMixMatch & 27.76 \\
w/ Re-sample & 27.22 \\
w/ LDAM-DRW & 30.22 \\
w/ DARP & 28.29 \\
w/ DARP + cRT & 30.13 \\
w/ CReST+ & 28.76 \\
w/ CReST+ + LA & 28.32 \\
w/ CoSSL & \textbf{31.31} \\ \bottomrule
\end{tabular}
}%
\resizebox{0.48\linewidth}{!}{%
\begin{tabular}{lcc}
\hline
CIFAR-100 & $N_1$=50 \\ \hline
FixMatch & 24.00 \\
w/ Re-sample & 25.06 \\
w/ LDAM-DRW & 23.30 \\
w/ DARP & 25.02 \\
w/ DARP + cRT & 24.55 \\
w/ CReST+ & 25.22 \\
w/ CReST+ + LA & 26.08 \\
w/ CoSSL & \textbf{28.42} \\ \bottomrule
\end{tabular}
}%
\caption{
Efficacy of CoSSL with less labeled data on CIFAR-100 with $\gamma=100$.
}
\label{tab:less}
\end{table}

\section{CoSSL pseudo-code} \label{sec:pseudocode}
We present the complete algorithm of our co-learning framework processing one batch of labeled and unlabeled images in algorithm \ref{alg}.

\begin{algorithm}[ht]
\caption{Co-learning of representation and classifier}
\begin{algorithmic}[1]  \label{alg}
    \STATE {\bfseries Input:} Labeled set $\mathcal{X} = \big\{(\textbf{x}_n, y_n): n \in (1,\ldots,N)\big\}$, unlabeled set $\mathcal{U} = \big\{\textbf{u}_m: m \in (1,\ldots,M)\big\}$, feature encoder $g$, classifier head in representation learning $h_r$, classifier head in classifier learning $h_c$, control parameter for fusion factor $\mu$, momentum coefficient $m$, batch size $B$, total number of training iterations $T$ \\
    \STATE $\xi^0=g$; $g^0=g$; $h_r^0=h_r$; $h_c^0=h_c$
    \FOR{$t = 0$ \TO $T-1$}
    \STATE \COMMENT{Sample labeled and unlabeled data for SSL} \\
    \STATE $\{\textbf{x}_i^t, y_i^t\}_{i=0}^{B-1}$ $\sim$ Random sampler($\mathcal{X}$) \\
    \STATE $\{\textbf{u}_i^t\}_{i=0}^{B-1}$ $\sim$ Random sampler($\mathcal{U}$) \\
    \STATE \COMMENT{Pseudo-labeling with EMA encoder and classifier} \\
    \STATE $\hat{y}_i^t$ = Pseudo-label($\xi^t$, $h_c^t$, $\textbf{u}_i^t$) $\forall i$
    \STATE \COMMENT{Apply TFE} \\
    \STATE $\{\Tilde{\textbf{z}}_i, \Tilde{y}_i\}_{i=0}^{B-1}$ = TFE($\mathcal{X}$, $\mathcal{U}$, $\xi^t$, $\mu$)
    \STATE \COMMENT{EMA update of the encoder} \\
    \STATE $\xi^{t+1} = m \xi^t + (1 - m) g^t$
    \STATE \COMMENT{Compute losses and update the model} \\
    \STATE $\mathcal{L}_x = \frac{1}{B} \sum_{i=1}^{B} \ell_{CE}(y_i^t, h_r^t(g^t(\textbf{x}_i^t))$ \\
    \STATE $\mathcal{L}_u = \frac{1}{B} \sum_{i=1}^{B} \ell_{CE}(\hat{y}_i^t, h_r^t(g^t(\textbf{u}_i^t))$ \\
    \STATE $\mathcal{L}_c = \frac{1}{B} \sum_{i=1}^{B} \ell_{CE}(\Tilde{y}_i^t, h_c^t(\Tilde{\textbf{z}}_i^t))$ \\
    \STATE $\mathcal{L} = \mathcal{L}_c + \mathcal{L}_x + \mathcal{L}_u$ \\
    \STATE $g^{t+1}$, $h_r^{t+1}$, $h_c^{t+1}$ = Update($g^t$, $h_r^t$, $h_c^t$) \\
    \ENDFOR
    \RETURN $\xi^{T}, 
    h_c^{T}$ \COMMENT{Model for evaluation} \\
\end{algorithmic}
\end{algorithm}

\section{Ablation study} \label{sec:ablation}
In this section, we provide more ablation results about different design choices of our method.

\myparagraph{Benefits of the co-learning framework.}
As argued in the main paper, we attribute the success of CoSSL to four aspects:
(1) Decoupling representation and classifier while coupling them closely (Table \ref{tab:cossl_jointvstwostage}).
(2) Classifier helps representation via pseudo-labeling rather using gradient directly (Table \ref{tab:cossl_sg}).
(3) Using the balanced classifier $h_{CL}$ for pseudo-label generation (Table \ref{tab:cossl_pl}).
(4) Using TFE for classifier learning (Table \ref{tab:cossl_tfe}).

\begin{table}[!ht]
    \centering
    \resizebox{\linewidth}{!}{%
    \begin{tabular}{lccccccc}
    \toprule
    \multicolumn{2}{l}{\multirow{2}{*}{\begin{tabular}[c]{@{}l@{}}Benefits of \\ decoupling\end{tabular}}} & \multicolumn{3}{c}{CIFAR-10} & \multicolumn{3}{c}{CIFAR-100} \\ \cmidrule(l){3-8} 
    \multicolumn{2}{l}{} & $\gamma$=50 & $\gamma$=100 & $\gamma$=150 & $\gamma$=20 & $\gamma$=50 & $\gamma$=100 \\ \midrule
    \multirow{3}{*}{Fix.} & - & 81.44 & 75.31 & 69.16 & 48.41 & 41.76 & 36.79 \\
     & two-stage & 82.93 & 78.51 & 73.52 & 49.95 & 44.11 & 39.54 \\
     & CoSSL & \textbf{86.42} & \textbf{82.60} & \textbf{80.24} & \textbf{52.76} & \textbf{47.04} & \textbf{42.09} \\ \midrule
    \multirow{3}{*}{ReMix.} & - & 82.57 & 76.94 & 73.30 & 50.76 & 43.51 & 38.48 \\
     & two-stage & 86.43 & 82.27 & 80.30 & 54.07 & 47.25 & 41.87 \\
     & CoSSL & \textbf{87.55} & \textbf{83.40} & \textbf{81.95} & \textbf{55.01} & \textbf{48.26} & \textbf{43.14} \\ \bottomrule
    \end{tabular}
    }%
    \caption{
    Both decoupled approaches (two-stage, CoSSL) show better results over the joint training. Particularly, our co-learning achieves the best performance across settings.
    }
    \label{tab:cossl_jointvstwostage}
\end{table}

\begin{table}
\centering
\resizebox{\linewidth}{!}{%
\begin{tabular}{llcccccc}
\toprule
\multicolumn{2}{l}{\multirow{2}{*}{\begin{tabular}[c]{@{}l@{}}Pseudo-label\\ generation\end{tabular}}} & \multicolumn{3}{c}{CIFAR-10} & \multicolumn{3}{c}{CIFAR-100} \\ \cmidrule(l){3-8} 
\multicolumn{2}{l}{} & $\gamma$=50 & $\gamma$=100 & $\gamma$=150 & $\gamma$=20 & $\gamma$=50 & $\gamma$=100 \\ \midrule
Fix. & $h_{SSL}$ & 85.48 & 81.20 & 78.23 & 52.24 & 45.90 & 40.43 \\
 & $h_{CL}$ & \textbf{86.42} & \textbf{82.60} & \textbf{80.24} & \textbf{52.76} & \textbf{47.04} & \textbf{42.09} \\ \midrule
ReMix. & $h_{SSL}$ & 86.90 & 82.88 & 80.22 & 54.39 & 47.81 & 42.09 \\
 & $h_{CL}$ & \textbf{87.55} & \textbf{83.40} & \textbf{81.95} & \textbf{55.01} & \textbf{48.26} & \textbf{43.14} \\ \bottomrule
\end{tabular}
}%
\caption{Benefits of using classifier learning module to generate pseudo-labels.
$h_{SSL}$ denotes the classifier from the representation learning module, $h_{CL}$ denotes the classifier from the classifier learning module.
}
\label{tab:cossl_pl}
\end{table}

\begin{table}
\centering
\resizebox{\linewidth}{!}{%
\begin{tabular}{llcccccc}
\toprule
\multicolumn{2}{l}{\multirow{2}{*}{Test Acc.}} & \multicolumn{3}{c}{CIFAR-10} & \multicolumn{3}{c}{CIFAR-100} \\ \cmidrule(l){3-8} 
\multicolumn{2}{l}{} & $\gamma$=50 & $\gamma$=100 & $\gamma$=150 & $\gamma$=20 & $\gamma$=50 & $\gamma$=100 \\ \midrule
Fix. & allow grad & \multicolumn{1}{l}{84.29} & \multicolumn{1}{l}{79.21} & \multicolumn{1}{l}{76.46} & \multicolumn{1}{l}{50.24} & \multicolumn{1}{l}{43.72} & \multicolumn{1}{l}{39.82} \\
 & CoSSL & \textbf{86.42} & \textbf{82.60} & \textbf{80.24} & \textbf{52.76} & \textbf{47.04} & \textbf{42.09} \\ \midrule
ReMix. & allow grad & \multicolumn{1}{l}{78.18} & \multicolumn{1}{l}{69.99} & \multicolumn{1}{l}{68.12} & \multicolumn{1}{l}{54.28} & \multicolumn{1}{l}{47.06} & \multicolumn{1}{l}{42.65} \\
 & CoSSL & \textbf{87.55} & \textbf{83.40} & \textbf{81.95} & \textbf{55.01} & \textbf{48.26} & \textbf{43.14} \\ \bottomrule
\end{tabular}
}%
\caption{Benefits of not updating the encoder from the gradient of the classifier module.}
\label{tab:cossl_sg}
\end{table}

\begin{table}
\centering
\resizebox{\linewidth}{!}{%
\begin{tabular}{lccccccc}
\toprule
\multirow{2}{*}{Test Acc.} & \multirow{2}{*}{Enhancement} & \multicolumn{3}{c}{CIFAR-10} & \multicolumn{3}{c}{CIFAR-100} \\ \cmidrule(l){3-8} 
\multicolumn{2}{l}{} & $\gamma$=50 & $\gamma$=100 & $\gamma$=150 & $\gamma$=20 & $\gamma$=50 & $\gamma$=100 \\ \midrule
\multirow{4}{*}{Fix.} & - & 84.24 & 80.27 & 77.22 & 51.40 & 45.39 & 41.33 \\
 & mixUp & 85.07 & 80.42 & 77.36 & 52.01 & 45.85 & 41.24 \\
 & MFW & 85.54 & 81.77 & 77.91 & 52.05 & 46.09 & 41.61 \\
 & TFE & \textbf{86.42} & \textbf{82.60} & \textbf{80.24} & \textbf{52.76} & \textbf{47.04} & \textbf{42.09} \\ \midrule
\multirow{4}{*}{ReMix.} & - & 87.06 & 82.24 & 79.53 & 54.58 & 47.84 & 42.60 \\
 & mixUp & 86.80 & 83.10 & 81.75 & \textbf{55.01} & 47.89 & 42.27 \\
 & MFW & 87.37 & \textbf{83.56} & 81.48 & 54.77 & 47.96 & 42.51 \\
 & TFE & \textbf{87.55} & 83.40 & \textbf{81.95} & \textbf{55.01} & \textbf{48.26} & \textbf{43.14} \\ \bottomrule
\end{tabular}
}%
\caption{Test accuracy of using different classifier learning methods in CoSSL.}
\label{tab:cossl_tfe}
\end{table}

\begin{figure}[!ht]
    \centering
    \begin{minipage}{0.48\linewidth}
        \includegraphics[width=\linewidth]{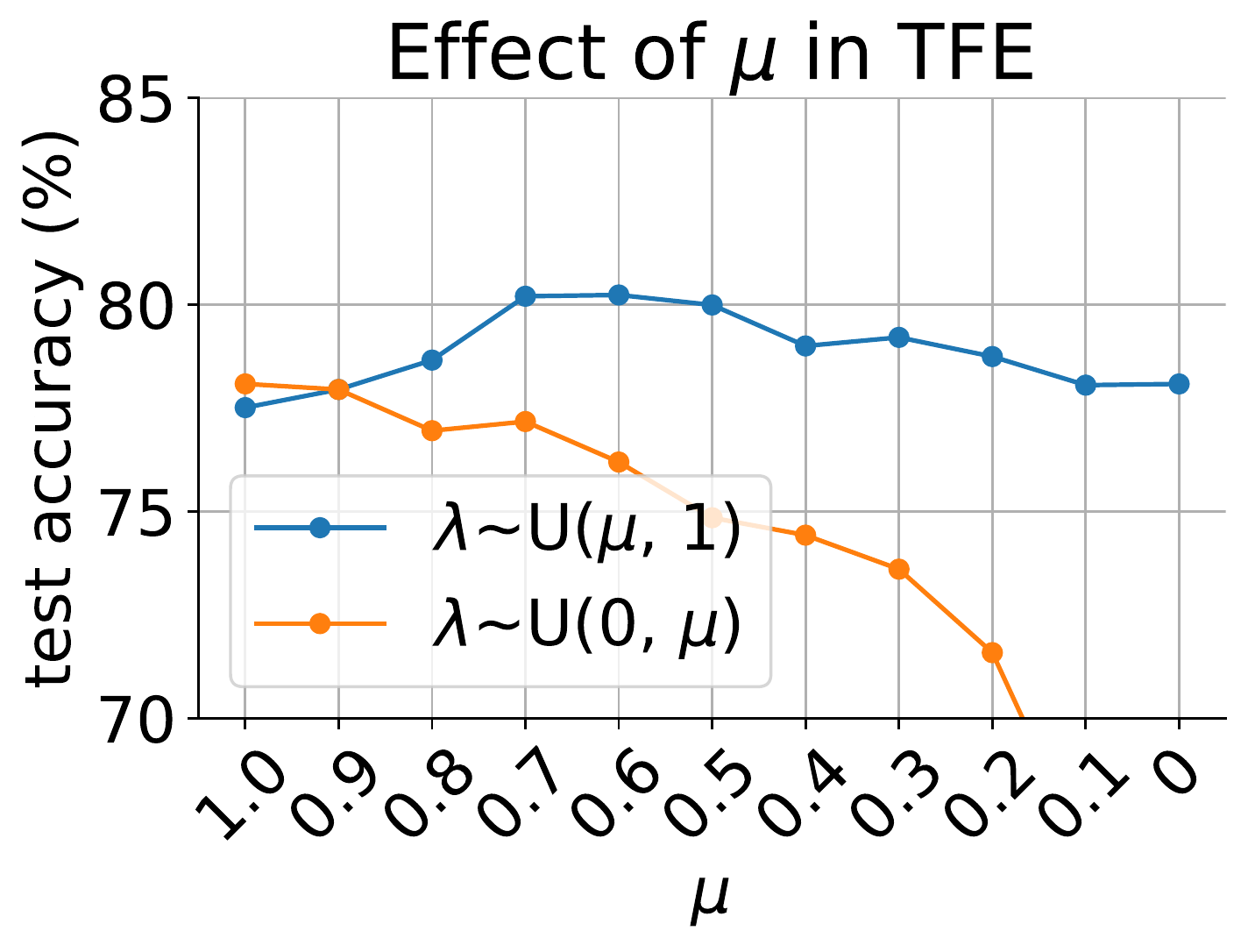}
    \end{minipage}
    \begin{minipage}{0.48\linewidth}
        \includegraphics[width=\linewidth]{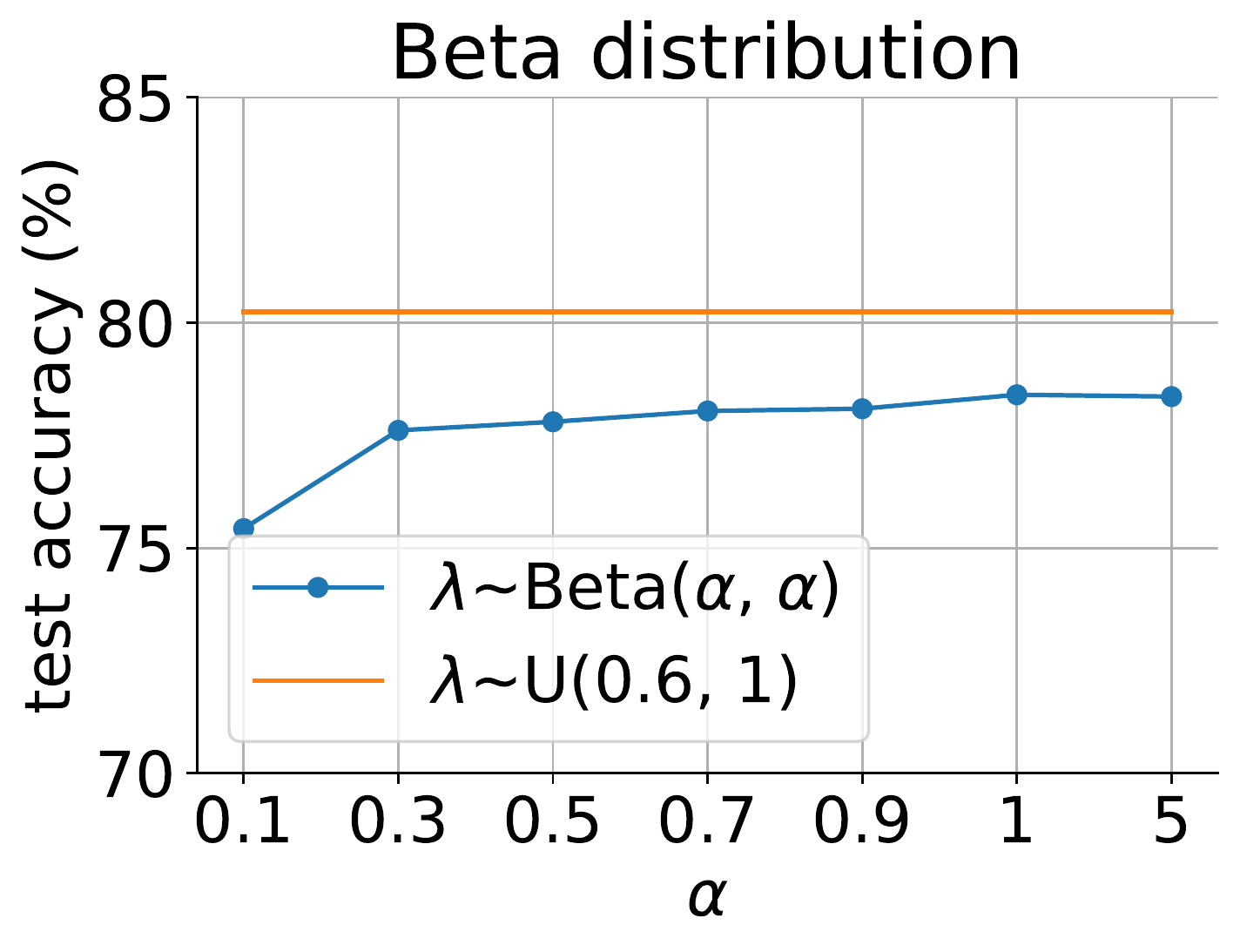}
    \end{minipage}
    \caption{
    \textbf{Left:}
    It is important to make $\lambda$ closer to 1 so that the blended feature is closer to the labeled feature, thus safer to share the label.
    Our method also shows good robustness within a wide range of $\mu$.
    \textbf{Right:}
    Comparison with beta distribution.
    }
    \label{fig:tfe_mu}
\end{figure}

\myparagraph{Sampling the fusion factor from a uniform distribution with lower bound.}
Here we study the effect of different $\mu$ from TFE Algorithm.
Since the fusion factor $\lambda$ is sampled from a uniform distribution between $\mu$ and 1, $\mu$ controls the regularization effect of feature blending.
A large $\mu$ indicates less regularization as the newly generated feature will be dominated by the labeled feature.
In the extreme cases, when $\mu=1$, TFE reduces to vanilla cRT as the unlabeled portion in the new feature is 0.
On the other hand, a small $\mu$ implies strong regularization as the new feature can potentially contain a large portion of unlabeled data while still using the same label.
As is shown by the blue curve in Fig. \ref{fig:tfe_mu} left, a $\mu$ with proper amount of regularization needs to be selected to maximize the model performance.
While $\mu = 0.6$ gives the best result ($80.24\%$), our model is quite robust within a large range of $\mu$.
Note that $\mu = 0.6$ is used as the default for all the results across datasets (CIFAR, ImageNet, and Food-101) in the main paper, which also indicates the robustness of our method.

Furthermore, as is compared in Figure \ref{fig:tfe_mu} left, sampling from the other half of the uniform distribution performs worse for all $\mu$ but the full range. 
Since the newly generated feature shares the class label with its labeled component, therefore, it is more beneficial to set $\lambda$ closer to 1 by sampling from a uniform distribution between $\mu$ and 1.
Moreover, the best uniform distribution with $\mu=0.6$ outperforms commonly used beta distribution as shown in Figure \ref{fig:tfe_mu} right.

\begin{figure}
\centering
\includegraphics[width=0.7\linewidth]{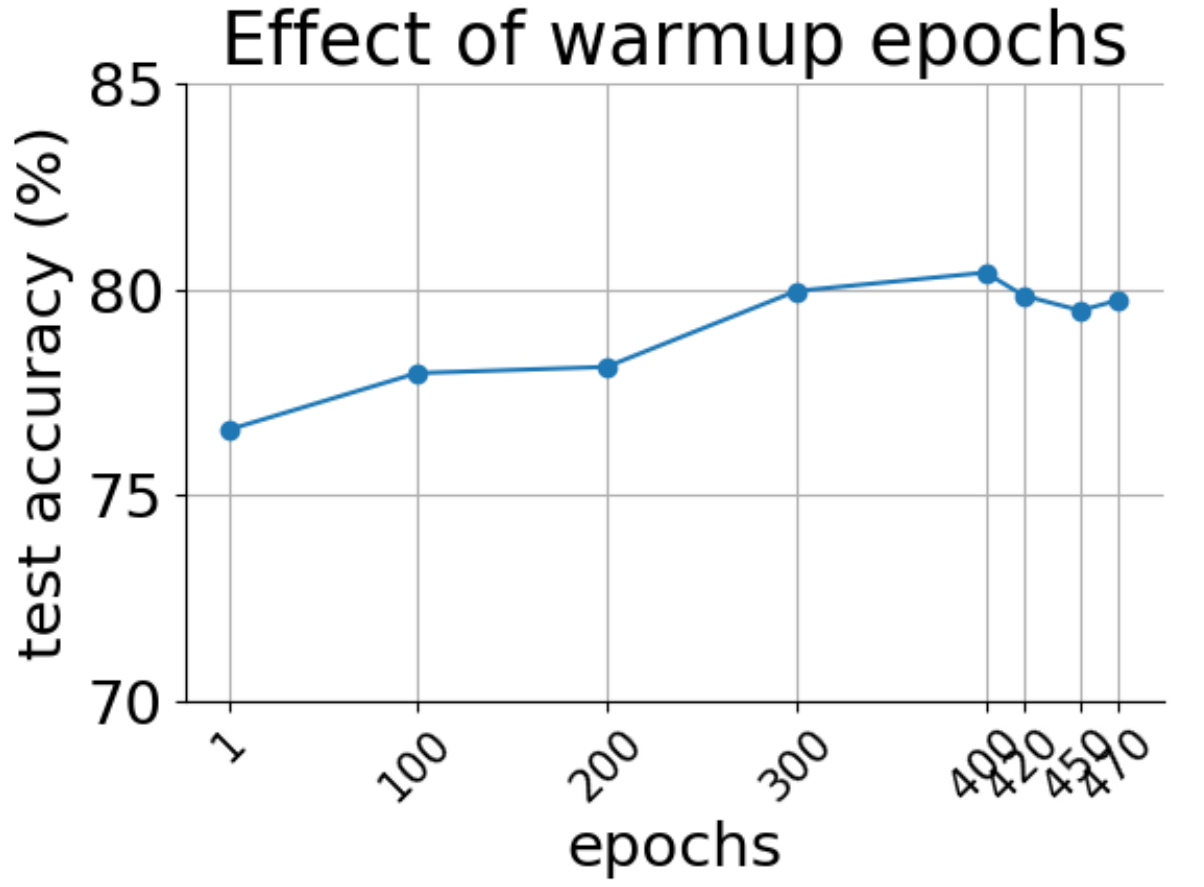}
\caption{
Effect of number of warm-up epochs for our method.
Enabling our co-learning at a later epoch is more beneficial.
}
\label{fig:ablation_warm}
\end{figure}

\myparagraph{Effect of the number of warm-up epochs.}
Here we study the effect of the number of warm-up epochs for representation learning.
Using a warm-up for representation learning can make the model enjoy both high precision of pseudo-labels in early training, and stronger class-rebalancing in late training.
Similar strategies are also widely used in many other works \cite{cao2019ldam,kim2020darp,wei2021crest}.
As shown in Fig. \ref{fig:ablation_warm}, warming up for longer than 300 epochs gives similar final results, and 400 epochs of warming-up achieves the best test accuracy, which corresponds to 80\% of the training time.
Our model shows good robustness in terms of the warming-up as we use a warm-up for the first 80\% of the training epochs for all experiments in the main paper and achieve good performance.

\myparagraph{Effect of the joint training.}
Here we extend the ablation study of the benefits of our co-learning framework compared with two-stage approaches.
Specifically, we compare our CoSSL with three variants of two-stage methods: vanilla cRT \cite{kang2019decouple}, cRT with mixUp, and our TFE.
For all three two-stage approaches, we first train a complete FixMatch for representation learning. 
Then, keeping the feature encoder fixed, the classification layer is reinitialized and trained for 20 epochs.
Table \ref{tab:ablate_coteach} summarizes the results.
In most cases, CoSSL outperforms two-stage methods, which demonstrates the benefits of the joint framework.
Moreover, TFE also fits particularly well for imbalanced SSL as TFE is in the top-two performing methods across different settings among two-stage methods.

\begin{table}[]
\resizebox{0.99\linewidth}{!}{%
\begin{tabular}{lcccccc}
\toprule
\multirow{2}{*}{Ablation} & \multicolumn{3}{c}{CIFAR-10} & \multicolumn{3}{c}{CIFAR-100} \\ \cmidrule(l{3pt}r{3pt}){2-4} \cmidrule(l{3pt}r{3pt}){5-7}
 & $\gamma$=50 & $\gamma$=100 & $\gamma$=150 & $\gamma$=20 & $\gamma$=50 & $\gamma$=100 \\ \cmidrule(l{3pt}r{3pt}){1-1} \cmidrule(l{3pt}r{3pt}){2-7}
FixMatch & 81.44 & 75.31 & 69.16 & 48.41 & 41.76 & 36.79 \\
+ cRT & 82.93 & 78.51 & 73.52 & 49.95 & 44.11 & 39.54 \\
+ cRT w/ mixUp & 86.16 & 81.94 & 77.50 & 51.08 & 43.74 & 39.18 \\
+ TFE & \textbf{86.83} & 81.94 & 77.93 & 52.88 & 45.37 & 40.79 \\ \cmidrule(l{3pt}r{3pt}){1-1} \cmidrule(l{3pt}r{3pt}){2-7}
FixMatch + CoSSL & 86.42 & \textbf{82.60} & \textbf{80.24} & \textbf{52.76} & \textbf{47.04} & \textbf{42.09} \\ \bottomrule
\end{tabular}
}%
\caption{
Classification accuracy (\%) of two-stage methods compared to our CoSSL.
The better performance demonstrates the effectiveness of our co-learning framework.
}
\label{tab:ablate_coteach}
\end{table}

\myparagraph{Class-imbalanced sampler during the classifier training.}
Here we study the effect of the class-imbalanced sampler in TFE under known shifted test distributions.
When the shifted distribution is known prior to the training, we can leverage this information to improve the performance at important classes by replacing the class-balanced sampler in TFE with a sampler following the target distribution.
Specifically, we train our models using class-imbalanced samplers with various imbalance ratios during the classifier training, and test them under three known shifted distributions.
We report classwise accuracies on CIFAR-10-LT with an imbalance ratio of 150 and use FixMatch as the base SSL method.

\begin{figure*}
\centering
\begin{minipage}{0.33\linewidth}
\includegraphics[width=1\linewidth]{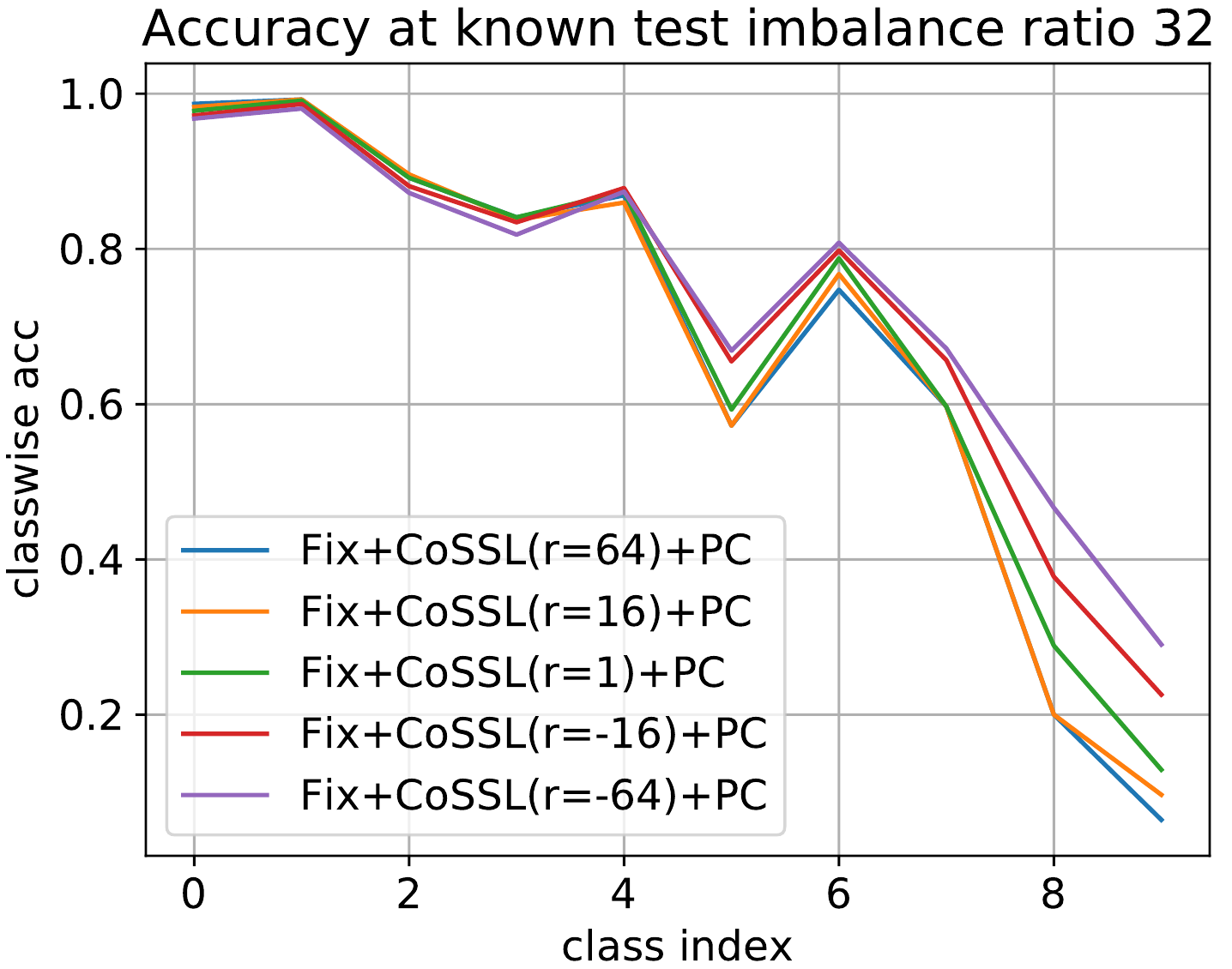}
\end{minipage}%
\begin{minipage}{0.33\linewidth}
\includegraphics[width=1\linewidth]{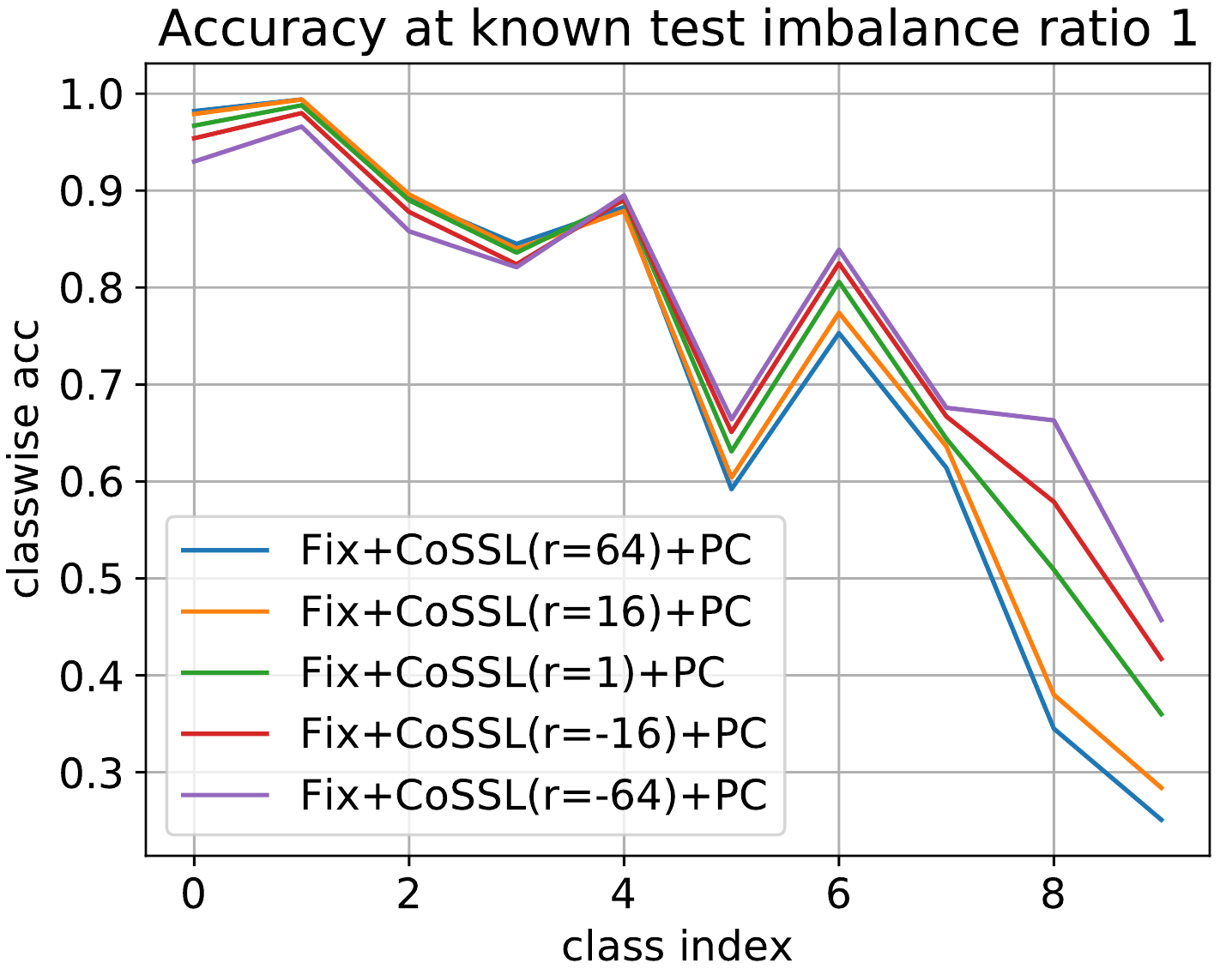}
\end{minipage}%
\begin{minipage}{0.33\linewidth}
\includegraphics[width=1\linewidth]{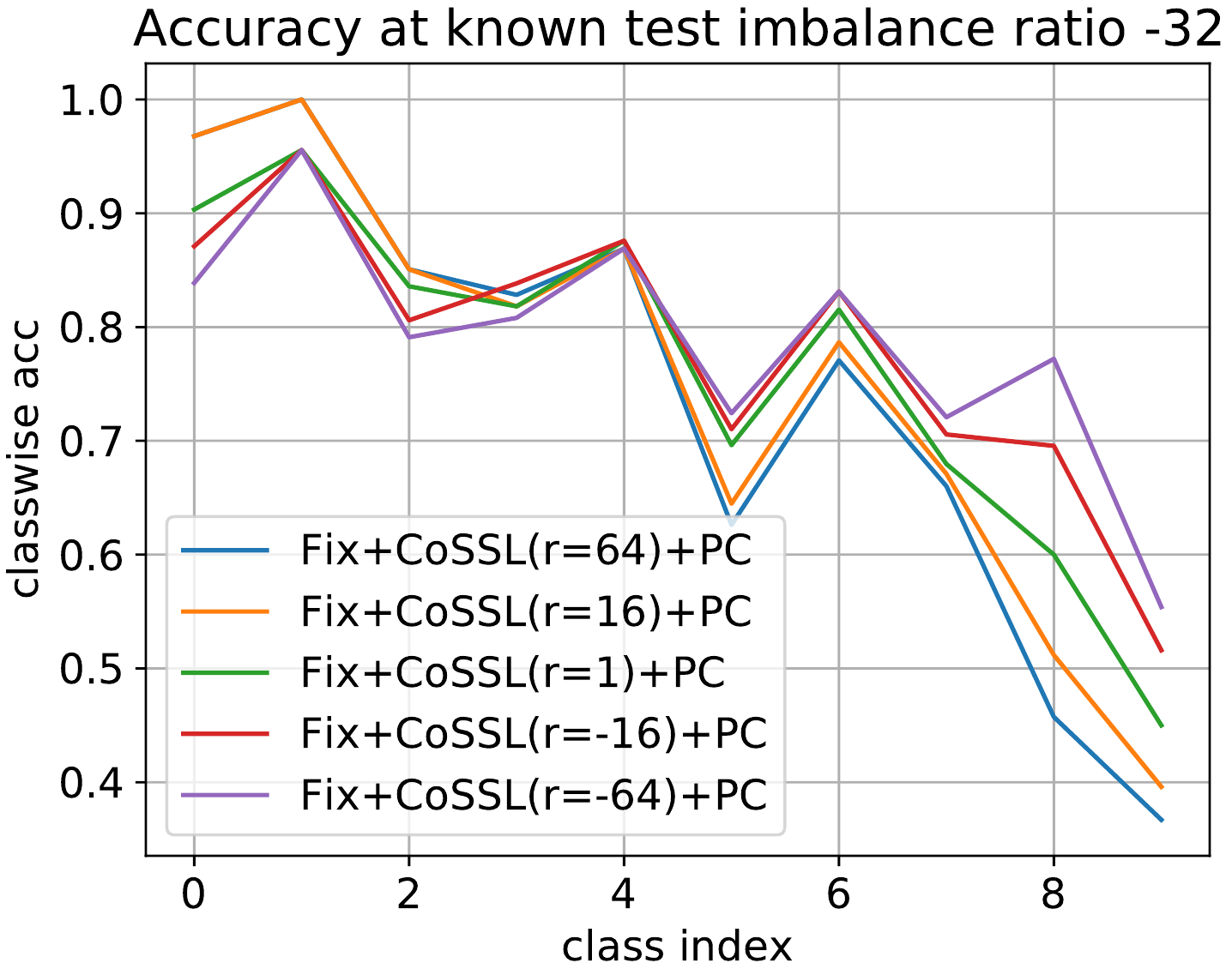}
\end{minipage}%
\caption{
Rather than only looking at average class accuracy, this figure shows classwise accuracies of CoSSL. In particular, we train with different class-imbalanced samplers (r = 64, 16, 1, -16, -64) under three known shifted evaluation settings (left ratio 32, middle 1, right -32). 
The train data has an imbalance ratio of 150.
In the case of test imbalance ratio of 32 (left figure), we can see that the class-imbalanced sampler has little effect on the head classes (classes 0, 1, etc) while having a strong influence on the tail classes (classes 9, 8, etc). This can be explained by the effect that a class-imbalanced sampler with a ratio of e.g. -64 will heavily oversample the tail classes and thus improve their performance overall. When the test imbalance ratio is further away (middle figure ratio 1, right figure ratio -32)  from the train data imbalance of 150 we can see a similar trend for the tail classes, however, the trend for the head classes is the opposite. Thus there is a clear trade-off between head and tail classes depending on which class-imbalanced sampler is used for training the classifier. Depending on the application scenario it might be thus interesting to not only look at average accuracies but more closely at this trade-off. 
}
\label{fig:classwise_acc}
\end{figure*}

Fig. \ref{fig:classwise_acc} shows the classwise accuracies at known test distributions with imbalance ratio $\gamma=32$, 1, and -32.
While the class-balanced sampler (r=1) gives reasonable performance across classes, using class-imbalanced sampler during the classifier training can make the model in favor of head or tail classes.
For example, when using a sampler with a large negative imbalance ratio -64, performance of tail classes can be improved further.
The trend of the head classes is, however, the opposite, which shows a clear trade-off. 
Therefore, depending on the target distribution, an imbalanced sampler favoring the important classes should be deployed to improve the overall performance.

Table. \ref{tab:skewed_c10_r150} summarizes the average class accuracy of CoSSL trained with different class-imbalanced sampler under known shifted distributions.
Replacing the class-balanced sampler in TFE with a sampler following the distribution of imbalance ratio 2 gives large improvement at positive test imbalance ratios and achieves the best numbers in most cases.

\begin{table*}[ht]
\resizebox{\textwidth}{!}{%
\centering
\begin{tabular}{lccccccccccccccccccccc}
\toprule
\multirow{2}{*}{Test imbalance ratio} & 512 & 256 & \textbf{150} & 128 & 64 & 32 & 16 & 8 & 4 & \multicolumn{1}{c|}{2} & \multicolumn{1}{c|}{1} & -2 & -4 & -8 & -16 & -32 & -64 & -128 & -256 & -512 & Mean \\  \cmidrule(l{3pt}r{3pt}){2-21} \cmidrule(l{3pt}r{3pt}){22-22}
 & \multicolumn{21}{c}{\cellcolor{blue!25}\textbf{Unknown test-time imbalance ratio}} \\ \cmidrule(l{3pt}r{3pt}){1-1} \cmidrule(l{3pt}r{3pt}){2-21} \cmidrule(l{3pt}r{3pt}){22-22}
Fix & 94.83 & 93.95 & 93.13 & 92.87 & 91.24 & 89.11 & 86.62 & 82.90 & 78.92 & \multicolumn{1}{c|}{73.58} & \multicolumn{1}{c|}{67.83} & 61.83 & 55.41 & 49.50 & 44.46 & 40.37 & 36.88 & 33.89 & 30.95 & 29.04 & 66.36 \\ \cmidrule(l{3pt}r{3pt}){1-1} \cmidrule(l{3pt}r{3pt}){2-21} \cmidrule(l{3pt}r{3pt}){22-22}
Fix + PC & 94.63 & 93.95 & 93.30 & 92.95 & 91.54 & 89.89 & 87.87 & 84.89 & 82.05 & \multicolumn{1}{c|}{77.97} & \multicolumn{1}{c|}{73.49} & 68.86 & 63.88 & 59.45 & 55.70 & 52.76 & 50.24 & 47.90 & 45.77 & 44.23 & 72.57 \\
Fix + vanilla cRT & 94.78 & 93.90 & 93.17 & 92.83 & 91.24 & 89.24 & 86.87 & 83.75 & 80.29 & \multicolumn{1}{c|}{75.54} & \multicolumn{1}{c|}{70.40} & 65.10 & 59.47 & 54.36 & 49.86 & 46.35 & 43.39 & 40.81 & 38.34 & 36.61 & 69.31 \\ \cmidrule(l{3pt}r{3pt}){1-1} \cmidrule(l{3pt}r{3pt}){2-21} \cmidrule(l{3pt}r{3pt}){22-22}
Fix + DARP & \textbf{95.14} & \textbf{94.46} & \textbf{93.73} & \textbf{93.50} & \textbf{92.18} & \textbf{90.12} & 87.70 & 84.39 & 81.03 & \multicolumn{1}{c|}{76.26} & \multicolumn{1}{c|}{71.15} & 66.12 & 60.99 & 56.10 & 52.28 & 48.84 & 45.75 & 43.25 & 40.79 & 39.17 & 70.65 \\
Fix + CReST+ & 94.18 & 93.39 & 92.74 & 92.45 & 91.05 & 89.04 & 86.70 & 83.52 & 80.20 & \multicolumn{1}{c|}{76.05} & \multicolumn{1}{c|}{71.75} & 67.28 & 62.76 & 58.73 & 55.68 & 52.89 & 50.47 & 48.49 & 46.61 & 45.54 & 71.98 \\
Fix + CoSSL & 91.73 & 91.13 & 90.90 & 90.60 & 89.85 & 89.07 & \textbf{87.95} & \textbf{86.24} & \textbf{84.60} & \multicolumn{1}{c|}{\textbf{82.61}} & \multicolumn{1}{c|}{\textbf{80.40}} & \textbf{78.39} & \textbf{76.03} & \textbf{74.19} & \textbf{73.21} & \textbf{72.49} & \textbf{71.43} & \textbf{70.64} & \textbf{70.02} & \textbf{69.71} & \textbf{81.06} \\ \cmidrule(l{3pt}r{3pt}){1-1} \cmidrule(l{3pt}r{3pt}){2-21} \cmidrule(l{3pt}r{3pt}){22-22}
 & \multicolumn{21}{c}{\cellcolor{blue!25}\textbf{Known test-time imbalance ratio}} \\ \cmidrule(l{3pt}r{3pt}){1-1} \cmidrule(l{3pt}r{3pt}){2-21} \cmidrule(l{3pt}r{3pt}){22-22}
Fix + PC & 94.98 & 94.00 & 93.13 & 92.83 & 91.16 & 89.24 & 87.03 & 84.00 & 81.03 & \multicolumn{1}{c|}{77.31} & \multicolumn{1}{c|}{73.49} & 70.10 & 66.79 & 64.21 & 62.69 & 61.89 & 62.41 & 63.26 & 64.80 & 66.50 & 77.04 \\
Fix + vanilla cRT & 95.14 & 94.32 & 93.39 & 93.25 & 91.35 & 89.24 & 86.73 & 83.45 & 79.85 & \multicolumn{1}{c|}{75.04} & \multicolumn{1}{c|}{70.40} & 65.76 & 60.65 & 56.67 & 53.81 & 52.04 & 51.07 & 51.09 & 49.98 & 51.60 & 72.24 \\
Fix + DARP + PC & 95.19 & \textbf{94.46} & 93.73 & 93.54 & 92.32 & 90.32 & 88.17 & 85.53 & 83.00 & \multicolumn{1}{c|}{79.96} & \multicolumn{1}{c|}{76.82} & 74.33 & 72.05 & 70.88 & 70.37 & 70.53 & 70.98 & 71.39 & 72.19 & 73.07 & 80.94 \\
Fix + CReST+ + PC & 94.48 & 93.44 & 92.74 & 92.49 & 91.09 & 89.17 & 87.20 & 84.75 & 82.60 & \multicolumn{1}{c|}{79.86} & \multicolumn{1}{c|}{77.74} & 76.09 & 74.41 & 74.03 & 74.40 & 75.40 & 76.38 & 77.22 & 78.66 & 80.29 & 82.62 \\
Fix + CoSSL + PC & 92.83 & 91.59 & 90.90 & 90.31 & 89.22 & 87.93 & 86.42 & 85.01 & 84.00 & \multicolumn{1}{c|}{82.57} & \multicolumn{1}{c|}{\textbf{82.00}} & \textbf{81.70} & \textbf{81.72} & \textbf{81.66} & \textbf{82.94} & \textbf{84.66} & \textbf{85.77} & \textbf{86.83} & \textbf{87.58} & \textbf{88.31} & \textbf{86.20} \\ \cmidrule(l{3pt}r{3pt}){1-1} \cmidrule(l{3pt}r{3pt}){2-21} \cmidrule(l{3pt}r{3pt}){22-22}
Fix + CoSSL(r=2) + PC & \textbf{95.24} & 94.27 & \textbf{93.95} & \textbf{93.67} & \textbf{92.51} & \textbf{91.07} & 89.23 & 86.89 & \textbf{85.05} & \multicolumn{1}{c|}{\textbf{82.64}} & \multicolumn{1}{c|}{80.87} & 79.95 & 78.89 & 78.56 & 79.49 & 81.32 & 82.74 & 84.23 & 85.59 & 87.16 & 86.17 \\
Fix + CoSSL(r=4) + PC & 94.88 & 94.04 & 93.47 & 93.16 & 92.10 & 90.61 & 88.81 & 86.41 & 84.62 & \multicolumn{1}{c|}{82.65} & \multicolumn{1}{c|}{80.87} & 79.89 & 78.96 & 78.70 & 79.77 & 81.39 & 82.03 & 83.60 & 84.94 & 86.56 & 85.87 \\
Fix + CoSSL(r=16) + PC & 95.04 & 94.18 & 93.77 & 93.29 & 92.21 & 90.68 & \textbf{89.29} & \textbf{86.98} & 84.91 & \multicolumn{1}{c|}{82.74} & \multicolumn{1}{c|}{80.98} & 80.00 & 78.61 & 78.70 & 79.80 & 81.22 & 82.52 & 83.77 & 84.90 & 86.41 & 86.00 \\
\bottomrule
\end{tabular}
}%
\caption{
Classification accuracy (\%) on CIFAR-10-LT with imbalance ratio $\gamma=150$.
We test different methods on top of FixMatch \cite{sohn2020fixmatch} for known and unknown shifted distributions.
Post-compensation (PC) \cite{hong2021disentangling} is deployed to utilize the information of the known test distribution. 
}
\vspace{-3pt}
\label{tab:skewed_c10_r150}
\end{table*}

\section{More evaluation at unknown and known shifted distributions} \label{sec:skewed}
Here we extend the evaluation of different methods at shifted test distributions in Section 4.4.
We report results at imbalance ratio $\gamma=100$ on CIFAR-10-LT and  $\gamma=20$, 50 and 100 for CIFAR100-LT.
All experiments are run with the same data split and the training protocol from Section 4.1.
We take FixMatch as the base SSL method and test post-compensation (PC) \cite{hong2021disentangling}, classifier retraining (cRT) \cite{kang2019decouple}, DARP \cite{kim2020darp}, CReST+ \cite{wei2021crest}, and our CoSSL over a family of shifted distributions.
As PC takes in target distribution $p_t$ to modify the logits at test time, we set $p_t$ as the uniform distribution and the used test distribution for unknown and known distributions, respectively.
For cRT, we reinitialize and train the classification layer for 20 epochs while keeping the feature encoder fixed after the representation learning.

Table \ref{tab:skewed_c10_r100}, \ref{tab:skewed_c100_r20}, \ref{tab:skewed_c100_r50}, \ref{tab:skewed_c100_r100} show the evaluation results.
For unknown distributions, while compromising at some positive ratios, CoSSL outperforms other methods by large margins at negative ratios, which leads to the overall higher mean accuracy across different settings.
This indicates that our method addresses the imbalance better than other methods that only perform well at distributions closer to the ones used during the training.
Similarly, we achieve a more balanced performance across various imbalance ratios for known distributions as well.

\begin{table*}[ht]
\resizebox{\textwidth}{!}{%
\centering
\begin{tabular}{lccccccccccccccccccccc}
\toprule
\multirow{2}{*}{Test imbalance ratio} & 512 & 256 & 128 & \textbf{100} & 64 & 32 & 16 & 8 & 4 & \multicolumn{1}{c|}{2} & \multicolumn{1}{c|}{1} & -2 & -4 & -8 & -16 & -32 & -64 & -128 & -256 & -512 & Mean \\ \cmidrule(l{3pt}r{3pt}){2-21} \cmidrule(l{3pt}r{3pt}){22-22}
 & \multicolumn{21}{c}{\cellcolor{blue!25}\textbf{Unknown test-time imbalance ratio}}  \\ \cmidrule(l{3pt}r{3pt}){1-1} \cmidrule(l{3pt}r{3pt}){2-21} \cmidrule(l{3pt}r{3pt}){22-22}
Fix & 95.24 & 94.69 & 93.62 & 93.14 & 92.36 & 90.74 & 88.59 & 85.79 & 83.14 & \multicolumn{1}{c|}{79.19} & \multicolumn{1}{c|}{75.23} & 71.48 & 67.74 & 64.14 & 61.44 & 59.31 & 57.36 & 55.87 & 54.27 & 53.11 & 75.82 \\
\cmidrule(l{3pt}r{3pt}){1-1} \cmidrule(l{3pt}r{3pt}){2-21} \cmidrule(l{3pt}r{3pt}){22-22}
Fix + PC & 94.83 & 94.46 & \textbf{93.71} & 93.38 & 92.70 & 91.59 & \textbf{89.84} & \textbf{87.69} & \textbf{85.92} & \multicolumn{1}{c|}{83.07} & \multicolumn{1}{c|}{80.43} & 78.07 & 75.81 & 73.75 & 72.45 & 71.64 & 70.95 & 70.30 & 69.56 & 69.36 & 81.98 \\
Fix + vanilla cRT & 95.64 & 94.92 & 93.79 & 93.22 & 92.40 & 90.87 & 88.76 & 86.02 & 83.76 & \multicolumn{1}{c|}{79.69} & \multicolumn{1}{c|}{76.27} & 73.08 & 69.67 & 66.90 & 64.69 & 63.04 & 61.74 & 60.65 & 59.45 & 58.93 & 77.67 \\ \cmidrule(l{3pt}r{3pt}){1-1} \cmidrule(l{3pt}r{3pt}){2-21} \cmidrule(l{3pt}r{3pt}){22-22}
Fix + DARP & 95.29 & 94.60 & 93.58 & 92.94 & 92.25 & 90.74 & 89.29 & 86.36 & 83.96 & \multicolumn{1}{c|}{80.18} & \multicolumn{1}{c|}{76.60} & 73.28 & 69.79 & 66.69 & 64.39 & 62.71 & 61.25 & 60.07 & 58.85 & 58.12 & 77.55 \\
Fix + CReST+ & \textbf{95.44} & \textbf{95.10} & 93.70 & \textbf{94.25} & \textbf{93.00} & \textbf{91.63} & 89.76 & 87.32 & 84.63 & \multicolumn{1}{c|}{80.91} & \multicolumn{1}{c|}{77.42} & 73.97 & 70.12 & 66.78 & 64.33 & 62.22 & 60.28 & 58.60 & 57.09 & 55.87 & 77.62 \\
Fix + CoSSL & 91.68 & 91.27 & 90.86 & 90.56 & 90.27 & 89.47 & 88.59 & 87.09 & 85.83 & \multicolumn{1}{c|}{\textbf{84.04}} & \multicolumn{1}{c|}{\textbf{82.52}} & \textbf{81.09} & \textbf{79.91} & \textbf{78.63} & \textbf{77.88} & \textbf{77.53} & \textbf{77.12} & \textbf{76.85} & \textbf{76.40} & \textbf{76.38} & \textbf{83.70} \\ \cmidrule(l{3pt}r{3pt}){1-1} \cmidrule(l{3pt}r{3pt}){2-21} \cmidrule(l{3pt}r{3pt}){22-22}
 & \multicolumn{21}{c}{\cellcolor{blue!25}\textbf{Known test-time imbalance ratio}} \\ \cmidrule(l{3pt}r{3pt}){1-1} \cmidrule(l{3pt}r{3pt}){2-21} \cmidrule(l{3pt}r{3pt}){22-22}
Fix + PC & 95.39 & 94.50 & 93.54 & 93.14 & 92.44 & 91.00 & 89.04 & 86.70 & 85.05 & \multicolumn{1}{c|}{82.52} & \multicolumn{1}{c|}{80.43} & 79.15 & 77.91 & 77.10 & 77.49 & 78.38 & 79.30 & 81.00 & 81.85 & 83.15 & 84.95 \\
Fix + vanilla cRT & 95.74 & 95.06 & 93.83 & 93.38 & 92.36 & 90.97 & 88.43 & 85.81 & 83.52 & \multicolumn{1}{c|}{79.46} & \multicolumn{1}{c|}{76.27} & 73.53 & 71.18 & 69.07 & 68.28 & 67.52 & 68.06 & 67.99 & 70.12 & 74.57 & 80.26 \\
Fix + DARP + PC & 95.29 & 94.50 & 93.54 & 92.94 & 92.21 & 90.94 & 89.59 & 87.37 & 85.89 & \multicolumn{1}{c|}{83.43} & \multicolumn{1}{c|}{81.57} & 80.53 & 79.58 & 79.12 & 79.97 & 81.35 & 82.22 & 83.56 & 84.80 & 85.96 & 86.22 \\
Fix + CReST+ + PC & \textbf{95.69} & \textbf{95.15} & \textbf{93.70} & \textbf{94.30} & \textbf{92.89} & \textbf{91.59} & \textbf{89.84} & \textbf{87.94} & \textbf{86.02} & \multicolumn{1}{c|}{83.86} & \multicolumn{1}{c|}{82.27} & 81.07 & 80.16 & 79.60 & 80.19 & 81.22 & 82.74 & 83.60 & 84.94 & 86.06 & 86.64 \\
Fix + CoSSL + PC & 92.83 & 91.87 & 91.07 & 90.56 & 90.08 & 88.98 & 87.48 & 85.99 & 84.82 & \multicolumn{1}{c|}{\textbf{83.97}} & \multicolumn{1}{c|}{\textbf{83.57}} & \textbf{83.43} & \textbf{83.91} & \textbf{84.30} & \textbf{85.23} & \textbf{86.59} & \textbf{88.39} & \textbf{89.35} & \textbf{89.38} & \textbf{89.87} & \textbf{87.58} \\
\bottomrule
\end{tabular}
}%
\caption{
Classification accuracy (\%) on CIFAR-10-LT with imbalance ratio $\gamma=100$.
We test different methods on top of FixMatch \cite{sohn2020fixmatch} for known and unknown test-time distributions.
Post-compensation (PC) \cite{hong2021disentangling} is deployed to utilize the information of the known test distribution.
}
\label{tab:skewed_c10_r100}
\end{table*}

\begin{table*}[ht]
\resizebox{\textwidth}{!}{%
\centering
\begin{tabular}{lccccccccccccccc}
\toprule
\multirow{2}{*}{Test imbalance ratio} & 64 & 32 & \textbf{20} & 16 & 8 & 4 & \multicolumn{1}{c|}{2} & \multicolumn{1}{c|}{1} & -2 & -4 & -8 & -16 & -32 & -64 & Mean \\ \cmidrule(l{3pt}r{3pt}){2-15} \cmidrule(l{3pt}r{3pt}){16-16}
 & \multicolumn{15}{c}{\cellcolor{blue!25}\textbf{Unknown test-time imbalance ratio}}  \\ \cmidrule(l{3pt}r{3pt}){1-1} \cmidrule(l{3pt}r{3pt}){2-15} \cmidrule(l{3pt}r{3pt}){16-16}
Fix & 69.07 & 67.18 & 65.25 & 64.56 & 61.02 & 57.15 & \multicolumn{1}{c|}{53.01} & \multicolumn{1}{c|}{48.30} & 43.89 & 39.53 & 35.27 & 31.23 & 28.12 & 25.60 & 49.23 \\
\cmidrule(l{3pt}r{3pt}){1-1} \cmidrule(l{3pt}r{3pt}){2-15} \cmidrule(l{3pt}r{3pt}){16-16}
Fix + PC & 67.41 & 65.92 & 64.43 & 63.87 & 60.92 & 57.90 & \multicolumn{1}{c|}{54.44} & \multicolumn{1}{c|}{50.46} & 46.66 & 42.98 & 39.30 & 35.74 & 33.07 & 30.72 & 50.99 \\
Fix + vanilla cRT & 66.64 & 65.23 & 63.66 & 63.15 & 60.39 & 57.92 & \multicolumn{1}{c|}{54.27} & \multicolumn{1}{c|}{50.41} & 46.94 & 43.35 & 39.80 & 36.58 & 34.15 & 32.12 & 51.04 \\ \cmidrule(l{3pt}r{3pt}){1-1} \cmidrule(l{3pt}r{3pt}){2-15} \cmidrule(l{3pt}r{3pt}){16-16}
Fix + DARP & \textbf{69.41} & \textbf{67.91} & \textbf{65.96} & \textbf{65.30} & \textbf{62.22} & 58.25 & \multicolumn{1}{c|}{54.59} & \multicolumn{1}{c|}{50.02} & 45.65 & 41.23 & 37.50 & 33.41 & 30.61 & 28.20 & 50.73 \\
Fix + CReST+ & 65.96 & 64.66 & 63.41 & 63.18 & 60.70 & 58.29 & \multicolumn{1}{c|}{55.16} & \multicolumn{1}{c|}{51.85} & 48.45 & 45.49 & 42.36 & 39.68 & 37.62 & 35.49 & 52.31 \\
Fix + CoSSL & 65.87 & 65.27 & 63.95 & 63.51 & 61.40 & \textbf{59.07} &  \multicolumn{1}{c|}{\textbf{56.21}} & \multicolumn{1}{c|}{\textbf{53.11}} & \textbf{50.08} & \textbf{47.06} & \textbf{44.31} & \textbf{41.30} & \textbf{39.60} & \textbf{37.84} & \textbf{53.47} \\ \cmidrule(l{3pt}r{3pt}){1-1} \cmidrule(l{3pt}r{3pt}){2-15} \cmidrule(l{3pt}r{3pt}){16-16}
 & \multicolumn{15}{c}{\cellcolor{blue!25}\textbf{Known test-time imbalance ratio}} \\ \cmidrule(l{3pt}r{3pt}){1-1} \cmidrule(l{3pt}r{3pt}){2-15} \cmidrule(l{3pt}r{3pt}){16-16}
Fix + PC & 69.28 & 67.08 & 65.25 & 64.62 & 60.97 & 57.81 & \multicolumn{1}{c|}{54.21} & \multicolumn{1}{c|}{50.46} & 47.14 & 44.30 & 41.60 & 39.68 & 39.49 & 39.42 & 52.95 \\
Fix + vanilla cRT & 67.92 & 66.14 & 64.39 & 63.54 & 60.44 & 57.84 & \multicolumn{1}{c|}{53.78} & \multicolumn{1}{c|}{50.41} & 47.29 & 44.39 & 42.15 & 40.37 & 39.03 & 38.61 & 52.59 \\
Fix + DARP + PC & \textbf{69.84} & \textbf{68.01} & \textbf{65.96} & \textbf{65.42} & \textbf{62.38} & \textbf{58.74} & \multicolumn{1}{c|}{\textbf{55.96}} & \multicolumn{1}{c|}{52.36} & 49.76 & 46.72 & 44.95 & 42.58 & 41.88 & 41.55 & 54.72 \\
Fix + CReST+ + PC & 66.51 & 64.80 & 63.41 & 63.24 & 60.82 & 58.55 & \multicolumn{1}{c|}{55.61} & \multicolumn{1}{c|}{\textbf{53.06}} & 50.91 & 49.18 & 47.83 & 46.76 & 46.97 & 46.67 & 55.31 \\
Fix + CoSSL + PC & 67.11 & 65.70 & 63.95 & 63.30 & 60.58 & 58.01 & \multicolumn{1}{c|}{55.00} & \multicolumn{1}{c|}{52.64} & \textbf{51.00} & \textbf{49.25} & \textbf{48.00} & \textbf{46.79} & \textbf{47.08} & \textbf{47.31} & \textbf{55.41} \\
\bottomrule
\end{tabular}
}%
\caption{
Classification accuracy (\%) on CIFAR-100-LT with imbalance ratio $\gamma=20$.
}
\label{tab:skewed_c100_r20}
\end{table*}

\begin{table*}[ht]
\resizebox{\textwidth}{!}{%
\centering
\begin{tabular}{lccccccccccccccc}
\toprule
\multirow{2}{*}{Test imbalance ratio} & 64 & \textbf{50} & 32 & 16 & 8 & 4 & \multicolumn{1}{c|}{2} & \multicolumn{1}{c|}{1} & -2 & -4 & -8 & -16 & -32 & -64 & Mean \\ \cmidrule(l{3pt}r{3pt}){2-15} \cmidrule(l{3pt}r{3pt}){16-16}
 & \multicolumn{15}{c}{\cellcolor{blue!25}\textbf{Unknown test-time imbalance ratio}}  \\ \cmidrule(l{3pt}r{3pt}){1-1} \cmidrule(l{3pt}r{3pt}){2-15} \cmidrule(l{3pt}r{3pt}){16-16}
Fix & 66.30 & 65.69 & 64.08 & 60.91 & 56.63 & 51.97 & \multicolumn{1}{c|}{47.15} & \multicolumn{1}{c|}{41.83} & 36.28 & 31.32 & 26.20 & 21.95 & 19.13 & 16.25 & 43.26 \\
\cmidrule(l{3pt}r{3pt}){1-1} \cmidrule(l{3pt}r{3pt}){2-15} \cmidrule(l{3pt}r{3pt}){16-16}
Fix + PC & 65.61 & 65.16 & 63.83 & 60.97 & 57.35 & 53.41 & \multicolumn{1}{c|}{49.09} & \multicolumn{1}{c|}{44.30} & 39.51 & 34.87 & 30.38 & 26.46 & 23.86 & 21.46 & 45.45 \\
Fix + vanilla cRT & 64.80 & 64.28 & 63.00 & 60.08 & 56.75 & 53.06 & \multicolumn{1}{c|}{48.79} & \multicolumn{1}{c|}{44.52} & 39.63 & 34.84 & 30.59 & 26.84 & 24.40 & 22.10 & 45.26 \\ \cmidrule(l{3pt}r{3pt}){1-1} \cmidrule(l{3pt}r{3pt}){2-15} \cmidrule(l{3pt}r{3pt}){16-16}
Fix + DARP & \textbf{66.51} & \textbf{65.97} & \textbf{64.37} & \textbf{61.12} & 57.16 & 52.40 & \multicolumn{1}{c|}{47.32} & \multicolumn{1}{c|}{42.14} & 36.65 & 31.33 & 26.54 & 22.31 & 19.21 & 16.51 & 43.54 \\
Fix + CReST+ & 65.19 & 64.56 & 63.29 & 60.70 & 57.20 & 53.35 & 48.94 & 44.66 & 39.84 & 35.45 & 31.36 & 27.83 & 25.09 & 22.35 & 45.70 \\
Fix + CoSSL & 63.91 & 63.72 & 62.78 & 60.29 & \textbf{57.54} & \textbf{54.30} & \multicolumn{1}{c|}{\textbf{50.99}} & \multicolumn{1}{c|}{\textbf{47.12}} & \textbf{42.69} & \textbf{38.19} & \textbf{34.16} & \textbf{30.64} & \textbf{27.98} & \textbf{25.81} & \textbf{47.15} \\ \cmidrule(l{3pt}r{3pt}){1-1} \cmidrule(l{3pt}r{3pt}){2-15} \cmidrule(l{3pt}r{3pt}){16-16}
 & \multicolumn{15}{c}{\cellcolor{blue!25}\textbf{Known test-time imbalance ratio}} \\ \cmidrule(l{3pt}r{3pt}){1-1} \cmidrule(l{3pt}r{3pt}){2-15} \cmidrule(l{3pt}r{3pt}){16-16}
Fix + PC & 66.25 & 65.69 & 64.15 & 60.94 & 57.01 & 52.91 & \multicolumn{1}{c|}{48.76} & \multicolumn{1}{c|}{44.30} & 40.15 & 36.08 & 32.92 & 29.98 & 28.66 & 27.90 & 46.84 \\
Fix + vanilla cRT & 66.17 & 65.81 & 63.97 & 61.06 & 57.11 & 53.06 & \multicolumn{1}{c|}{48.76} & \multicolumn{1}{c|}{44.52} & 40.41 & 36.35 & 32.99 & 30.99 & 29.82 & 29.61 & 47.19 \\
Fix + DARP + PC & \textbf{66.55} & \textbf{65.97} & \textbf{64.55} & \textbf{61.27} & \textbf{57.52} & 53.19 & \multicolumn{1}{c|}{48.77} & \multicolumn{1}{c|}{44.76} & 40.71 & 36.72 & 33.66 & 30.90 & 29.57 & 28.50 & 47.33 \\
Fix + CReST+ + PC & 65.19 & 64.56 & 63.10 & 60.62 & 57.37 & \textbf{53.86} & \multicolumn{1}{c|}{49.85} & \multicolumn{1}{c|}{46.62} & 42.69 & 39.75 & 37.45 & 35.29 & 34.01 & 33.06 & 48.82 \\
Fix + CoSSL + PC & 64.29 & 63.72 & 62.35 & 59.48 & 56.32 & 53.46 & \multicolumn{1}{c|}{\textbf{50.39}} & \multicolumn{1}{c|}{\textbf{47.29}} & \textbf{44.54} & \textbf{41.65} & \textbf{39.51} & \textbf{37.56} & \textbf{37.69} & \textbf{38.18} & \textbf{49.74} \\
\bottomrule
\end{tabular}
}%
\caption{
Classification accuracy (\%) on CIFAR-100-LT with imbalance ratio $\gamma=50$.
}
\label{tab:skewed_c100_r50}
\end{table*}

\begin{table*}[ht]
\resizebox{\textwidth}{!}{%
\centering
\begin{tabular}{lccccccccccccccc}
\toprule
\multirow{2}{*}{Test imbalance ratio} & \textbf{100} & 64 & 32 & 16 & 8 & 4 & \multicolumn{1}{c|}{2} & \multicolumn{1}{c|}{1} & -2 & -4 & -8 & -16 & -32 & -64 & Mean \\ \cmidrule(l{3pt}r{3pt}){2-15} \cmidrule(l{3pt}r{3pt}){16-16}
 & \multicolumn{15}{c}{\cellcolor{blue!25}\textbf{Unknown test-time imbalance ratio}}  \\ \cmidrule(l{3pt}r{3pt}){1-1} \cmidrule(l{3pt}r{3pt}){2-15} \cmidrule(l{3pt}r{3pt}){16-16}
Fix & \textbf{67.25} & \textbf{65.49} & \textbf{62.42} & 58.61 & 53.70 & 48.42 & \multicolumn{1}{c|}{42.87} & \multicolumn{1}{c|}{37.09} & 31.17 & 25.78 & 20.67 & 16.45 & 13.18 & 10.24 & 39.52 \\
\cmidrule(l{3pt}r{3pt}){1-1} \cmidrule(l{3pt}r{3pt}){2-15} \cmidrule(l{3pt}r{3pt}){16-16}
Fix + PC & 66.26 & 64.72 & 62.06 & 58.52 & 54.30 & 49.55 & \multicolumn{1}{c|}{44.59} & \multicolumn{1}{c|}{39.22} & 33.66 & 28.52 & 23.47 & 19.35 & 16.14 & 13.31 & 40.98 \\
Fix + vanilla cRT & 65.18 & 63.78 & 61.26 & 58.05 & 54.04 & 49.57 & \multicolumn{1}{c|}{44.61} & \multicolumn{1}{c|}{39.73} & 33.97 & 28.78 & 23.93 & 20.24 & 17.08 & 14.29 & 41.04 \\ \cmidrule(l{3pt}r{3pt}){1-1} \cmidrule(l{3pt}r{3pt}){2-15} \cmidrule(l{3pt}r{3pt}){16-16}
Fix + DARP & 66.21 & 64.63 & 61.91 & 58.20 & 53.61 & 48.42 & \multicolumn{1}{c|}{43.09} & \multicolumn{1}{c|}{37.44} & 31.50 & 26.34 & 21.67 & 17.41 & 13.97 & 11.18 & 39.68 \\
Fix + CReST+ & 65.65 & 64.38 & 61.88 & \textbf{58.94} & 54.62 & 49.93 & \multicolumn{1}{c|}{44.81} & \multicolumn{1}{c|}{39.60} & 34.31 & 29.58 & 24.38 & 20.01 & 16.68 & 13.48 & 41.30 \\
Fix + CoSSL & 64.15 & 62.93 & 61.23 & 58.47 & \textbf{54.95} & \textbf{51.17} & \multicolumn{1}{c|}{\textbf{46.58}} & \multicolumn{1}{c|}{\textbf{42.22}} & \textbf{36.97} & \textbf{32.17} & \textbf{27.69} & \textbf{24.25} & \textbf{20.83} & \textbf{17.83} & \textbf{42.96} \\ \cmidrule(l{3pt}r{3pt}){1-1} \cmidrule(l{3pt}r{3pt}){2-15} \cmidrule(l{3pt}r{3pt}){16-16}
 & \multicolumn{15}{c}{\cellcolor{blue!25}\textbf{Known test-time imbalance ratio}} \\ \cmidrule(l{3pt}r{3pt}){1-1} \cmidrule(l{3pt}r{3pt}){2-15} \cmidrule(l{3pt}r{3pt}){16-16}
Fix + PC & \textbf{67.25} & \textbf{65.44} & \textbf{62.42} & 58.76 & 54.06 & 49.29 & \multicolumn{1}{c|}{44.38} & \multicolumn{1}{c|}{39.22} & 33.97 & 29.53 & 25.13 & 22.16 & 19.89 & 18.05 & 42.11 \\
Fix + vanilla cRT & 66.73 & 64.76 & 62.02 & 58.14 & 53.92 & 49.31 & \multicolumn{1}{c|}{44.32} & \multicolumn{1}{c|}{39.73} & 34.37 & 29.68 & 25.34 & 22.51 & 20.32 & 19.03 & 42.16 \\
Fix + DARP + PC & 66.21 & 64.72 & 62.06 & 58.35 & 53.87 & 49.48 & \multicolumn{1}{c|}{44.52} & \multicolumn{1}{c|}{40.00} & 35.08 & 30.91 & 27.14 & 23.89 & 21.91 & 20.01 & 42.73 \\
Fix + CReST+ + PC & 65.65 & 64.29 & 61.81 & \textbf{59.00} & \textbf{55.26} & \textbf{50.71} & \multicolumn{1}{c|}{\textbf{45.95}} & \multicolumn{1}{c|}{41.56} & 37.09 & 33.25 & 29.68 & 27.08 & 25.49 & 24.32 & 44.37 \\
Fix + CoSSL + PC & 64.15 & 62.37 & 60.14 & 57.45 & 53.54 & 49.89 & \multicolumn{1}{c|}{45.52} & \multicolumn{1}{c|}{\textbf{42.27}} & \textbf{38.25} & \textbf{34.99} & \textbf{32.63} & \textbf{30.99} & \textbf{29.89} & \textbf{28.37} & \textbf{45.03} \\
\bottomrule
\end{tabular}
}%
\caption{
Classification accuracy (\%) on CIFAR-100-LT with imbalance ratio $\gamma=100$.
}
\label{tab:skewed_c100_r100}
\end{table*}

\end{document}